\documentclass[letterpaper]{article} 
\usepackage[preprint]{aaai2027}  
\usepackage[hyphens]{url}  
\usepackage{graphicx} 
\usepackage{natbib}  
\usepackage{caption} 
\usepackage{algorithm}
\usepackage{algorithmic}

\usepackage{newfloat}
\usepackage{listings}
\DeclareCaptionStyle{ruled}{labelfont=normalfont,labelsep=colon,strut=off} 
\floatstyle{ruled}
\newfloat{listing}{tb}{lst}{}
\floatname{listing}{Listing}

\usepackage{booktabs}
\usepackage{colortbl}
\usepackage{amsmath}
\usepackage{amssymb}

\newcommand{\backtrace}{\textsc{Backtrace}}
\newcommand{\bench}{\textsc{BackroomBench}}
\newtheorem{definition}{Definition}

\title{Skill Use or Skill Theater? Evaluating the Reasoning Backroom in Skill-Augmented Language Agents}
\author{
    Jinwei Hu\textsuperscript{\rm 1}, Yi Qi\textsuperscript{\rm 2}, Xinmiao Huang\textsuperscript{\rm 1}, Youcheng Sun\textsuperscript{\rm 3}, Yi Dong\textsuperscript{\rm 1}, Xiaowei Huang\textsuperscript{\rm 1}
}
\affiliations{
    \textsuperscript{\rm 1}School of Computer Science and Informatics, University of Liverpool, UK\\
    \textsuperscript{\rm 2}Department of Computer Science, University of Leeds, UK\\
    \textsuperscript{\rm 3}Department of Computer Science, Mohamed bin Zayed University of Artificial Intelligence, UAE
    
}

\begin{document}

\maketitle

\begin{abstract}
Reusable skills are becoming a standard interface for extending language agents with task procedures. Yet evaluators usually infer skill use from visible reasoning or the agent's own attribution. These signals show what the agent appears to use, not whether the skill changed its decision. We ask whether skill-augmented agents exhibit a \textbf{Reasoning Backroom}, a systematic gap between stated skill use and intervention-measured influence. We introduce \backtrace{}, an evaluation framework that pairs each skill-conditioned answer with a matched no-skill counterfactual, intervenes on skill meaning, wording, identity, content, and assignment, and elicits attribution only after the answer is committed. We instantiate the framework as \bench{}, a verified testbed spanning controlled logic and competition mathematics, multiple skill conditions, single-agent and multi-agent settings, and diverse model families. Our evaluation reveals a pervasive provenance failure. Across models and domains, stated skill use often remains stable while causal reliance and signed utility vary, producing both silent uptake and performative use. Behavioral effects follow procedural content more reliably than displayed skill identity, whereas stated attributions respond strongly to artifact availability. Observational detectors based on direct skill-use claims, text mentions, trace similarity, and an LLM judge do not identify which decisions actually depend on the skill. In multi-agent systems, skill influence can survive communication even after its source is lost, while no-skill teams still name skills and sources that were never supplied. These findings establish the Reasoning Backroom as a general AI provenance problem whose audit requires intervention.

\end{abstract}

\section{Introduction}

Reusable skills package instructions, procedures, and verification guidance that LLM agents can invoke when needed. They have become a practical interface for extending agents across software, cybersecurity, embodied, and reasoning tasks \citep{anthropicskills,Hu_Dong_Sun_Huang_2026,li2026skillsbench}. These skill-augmented agents produce not only answers but also reasoning traces, citations to retrieved procedures, and statements about which skills they believe they used. In practice, these observable signals form a \emph{reasoning front room} through which developers and evaluators can monitor or test how an answer was produced \cite{wang2026agenttracestrustsurvey,hu2026skillfuzz}. Existing benchmarks reinforce this view by asking whether skills improve performance, are retrieved appropriately, or can be generated from experience \citep{cho2026skillret,liu2026agenticskillsworkwild,zhou2026skillgenbench}. Yet a central assumption behind this interface remains untested. Does the front room faithfully reveal which skill influenced the decision, or do language agents exhibit a \textbf{Reasoning Backroom}?

The backroom can take several forms that look similar from the outside. A skill may change an answer but remain unacknowledged; it may be cited even though removing it changes nothing; or it may affect the decision in a way that is beneficial on some instances and harmful on others, despite an identical claim of use. Behavior may follow procedural content even when claimed use does not identify it. In a multi-agent system (MAS), the problem becomes harder as skill-shaped information passes through specialists, verifiers, and aggregators, allowing influence to survive after its source becomes ambiguous or reassigned. Neither final accuracy nor a plausible trace distinguishes these cases, and prior studies show that visible reasoning can omit or mislead behaviorally relevant inputs \citep{turpin2023language,paul2024making,hao-etal-2026-reasoning,arcuschin2025chainofthought,hu-etal-2026-lying}. The evaluation problem is therefore not simply whether a skill is present or mentioned, but whether its stated role matches its causal effect on the decision.

We investigate this question with \backtrace{}, a framework for evaluating \emph{causal skill attribution} without claiming direct access to a model's internal mechanism. \backtrace{} commits an answer before eliciting attribution, pairs each skill-conditioned run with a matched no-skill counterfactual, and intervenes on semantics, rules, names, content, assignment, and communication. These comparisons separate whether a skill changes the answer, whether that change helps or harms, and whether the system's stated skill use matches that dependence. We instantiate the framework as \bench{} based on Logic and MATH problems \cite{saparov2023language, hendrycks2021measuring}, seven frozen skill conditions, and matched single-agent and role-structured evaluations across twelve model variants. Logic provides rule chains and answers known by construction; Mathematics tests transfer to a natural domain.

Our audit finds that the Reasoning Backroom is pervasive rather than an isolated model failure. Across the tested models, interventions, reasoning depths, domains, and organizations, observable claims often remain stable while causal reliance and signed utility vary. Agents silently absorb some skills, confidently cite inert ones, and follow procedural content without identifying the active component. In MAS, the separation widens as influence survives communication after its source is lost, agents misidentify which teammate mattered, and no-skill teams name skills and sources that were never supplied. These mismatches persist under natural Mathematics tasks and are not resolved by detectors built from visible traces or explicit skill-use claims. An apparently coherent front room can therefore misstate whether, how, and through whom an external capability shaped a decision. The Reasoning Backroom is a general AI provenance problem whose causal claims require intervention.

Our contributions are summarized as below:
\begin{itemize}
    \item We formalize the Reasoning Backroom as an evaluation problem and characterize its individual and distributed manifestations, such as silent uptake, performative use, propagation, laundering, and contribution mismatch.
    \item We introduce \backtrace{} and \bench{}, combining controlled skill interventions with post-decision attribution and paired single-agent/MAS configuration.
    \item We establish twelve-model Logic and six-model cross-domain evidence that observable reasoning and stated skill use do not faithfully identify the magnitude, direction, or source of skill influence without intervention.
\end{itemize}

\section{Related Work}

\paragraph{The Rise of Agent Skills.}
Large language model agents increasingly extend their competence through external, reusable artifacts rather than parameter updates alone. Early tool-learning systems taught models to decide which APIs to invoke and how to compose calls for complex instructions \citep{schick2023toolformer,qin2024toolllm}. A parallel line of work moved procedural knowledge itself outside model weights. Reflexion retains linguistic feedback in episodic memory, ExpeL distills experience into reusable natural-language insights, AutoGuide extracts context-aware guidelines from offline trajectories, and TAPAS synthesizes executable programs that adapt an agent to changing environments \citep{shinn2023reflexion,zhao2024expel,fu2024autoguide,Hu_Dong_Sun_Huang_2026}. Although these artifacts differ in representation, they establish a common design pattern in which an agent stores acquired procedures and selectively reuses them on later tasks. Contemporary platforms implement this pattern as named skills that package instructions, applicability conditions, scripts, and supporting resources for inference-time invocation \citep{anthropicskills}. A skill therefore occupies a distinct role. A tool defines an available action, whereas a skill specifies procedural knowledge for deciding when and how that action should be used. Unlike an ephemeral prompt, a skill also has a stable identity and can be retrieved, revised, shared, and transferred between agents, making skills a practical unit of agent capability \cite{li2026skillsbench}.

\paragraph{Agent Evaluation.}
Agent evaluation has expanded with this ecosystem. General benchmarks measure end-to-end task success across heterogeneous interactive environments, enterprise knowledge work, and open-ended computer use \citep{liu2024agentbench,drouin2024workarena,xie2024osworld}. Tool-centered evaluations further test whether models can retrieve and execute APIs, benefit from tools and feedback over multiple turns, or complete the component stages of tool use such as planning, retrieval, reasoning, and review \citep{qin2024toolllm,wang2024mint,chen2024teval}. Recent skill-specific benchmarks ask whether supplied skill files improve outcomes and whether agents retrieve the appropriate artifact from a library \citep{li2026skillsbench,cho2026skillret}. Together, these evaluations cover realistic tasks and fine-grained agent stages, but their evidence centers on capability: task completion shows success, component scores locate failures, and retrieval confirms availability or selection. None establishes whether an answer counterfactually depended on skill content, whether that influence helped or harmed, or whether stated use matched it. We address this gap by testing the causal influence of external artifacts as part of execution provenance \citep{wang2026agenttracestrustsurvey,hu2026responsibleagenticairequires}. Unlike CoT faithfulness, which audits whether a generated rationale reflects the prediction process, skill provenance audits whether an external reusable artifact changed the decision and whether its identity and source are recovered.

\section{Problem Formulation}

\subsection{Skill-Augmented Decision Process}

Let $\mathcal{X}$ be the instance space and let $x\in\mathcal{X}$ have a verifier-approved target $y^*$. We audit a frozen skill family $s$ under conditions indexed by $v\in\mathcal{V}$. The artifact supplied in condition $v$ is $s_v$, with $s_{\varnothing}=\varnothing$ denoting the no-skill condition. Let $\pi=(\pi_{\mathrm{ans}},\pi_{\mathrm{attr}})$ denote the frozen agent system, where $\pi_{\mathrm{ans}}$ is its answer channel and $\pi_{\mathrm{attr}}$ is its post-decision attribution channel. The execution context $\omega$ collects all non-target factors held fixed within a comparison, including the model, prompt frame, decoding policy, system organization, and random state when it is controllable. The answer channel first produces a visible reasoning trace $\tau_v$ and free-form answer $\hat y_v$,
{\small
\begin{equation}
(\tau_v,\hat y_v)=\pi_{\mathrm{ans}}(x,s_v;\omega).
\end{equation}
}
Only after the answer is committed does the agent produce a structured attribution response
{\small
\begin{equation}
q_v=\pi_{\mathrm{attr}}(x,s_v,\tau_v,\hat y_v;\omega).
\end{equation}
}
This ordering prevents the attribution request from affecting the decision it is meant to describe. Let $\nu$ be the deterministic domain normalizer, let $d_v(x;\omega)=\nu(\hat y_v)$ be the normalized decision, and let $d^*=\nu(y^*)$ be the normalized target. We write $d_v$ when the instance and execution context are clear and suppress $\omega$ from subsequent per-instance quantities. A validated extraction function $g$ maps $q_v$ to
{\small
\begin{equation}
g(q_v)=\bigl(a_v(s,x),\hat p_v(s,x)\bigr),
\end{equation}
}
where $a_v\in\{0,1\}$ records whether the system claims to have used the provided skill and $\hat p_v$ records its claimed provenance. Provenance is the skill identity in a single-agent system and the skill-source pair in a multi-agent system.

\subsection{The Reasoning Front Room}

\begin{definition}[Reasoning front room]
\label{def:frontroom}
The reasoning front room for condition $v$ is the observable record
{\small
\begin{equation}
F_v(x)=\bigl(\tau_v,d_v,q_v\bigr).
\end{equation}
}
\end{definition}
The front room contains all signals available to an observation-only audit. It shows the answer, exposed reasoning, and credited skill, but not whether the answer depended on that skill. Accuracy is likewise insufficient because it evaluates the outcome rather than its cause.

\subsection{The Intervention-Defined Reasoning Backroom}

We identify answer-level dependence by coupling the skill-conditioned decision $d_v(x;\omega)$ with the no-skill decision $d_{\varnothing}(x;\omega)$ under the same execution context. For any proposition $P$, let $\mathbb{1}[P]$ equal one when $P$ is true and zero otherwise. The intervention defines causal reliance and signed utility as
{\small
\begin{equation}
\label{def:reliance}
\begin{aligned}
r_v(s,x)&=\mathbb{1}\!\left[d_v\neq d_{\varnothing}\right],\\
u_v(s,x)&=\mathbb{1}[d_v=d^*]
-\mathbb{1}[d_{\varnothing}=d^*].
\end{aligned}
\end{equation}
}
Reliance records decision change. Signed utility records whether it helps or harms and is zero otherwise. This answer-level measure is conservative. When $r_v=0$, deletion preserves the normalized answer, but the skill may still alter latent computation.

\begin{definition}[Reasoning backroom]
\label{def:backroom}
The reasoning backroom is the intervention-defined state
{\small
\begin{equation}
B_v(x)=\bigl(r_v(s,x),u_v(s,x)\bigr).
\end{equation}
}
\end{definition}
This definition does not claim access to the model's private computation. It isolates a behavioral fact unavailable from a single visible trace. A Reasoning Backroom failure occurs when the front-room claim disagrees with intervention-defined dependence,
{\small
\begin{equation}
m_v(s,x)=\mathbb{1}\!\left[a_v(s,x)\neq r_v(s,x)\right].
\end{equation}
}
The pair $(r_v,a_v)$ induces four instance-level states. $(1,1)$ is \emph{faithful explicit use}, and $(0,0)$ is \emph{correct rejection}. $(1,0)$ is \textbf{silent uptake}, where the artifact changes the answer without being claimed. $(0,1)$ is \textbf{performative use}, where the artifact is claimed despite having no answer-level effect. The utility variable $u_v$ further establishes whether any measured dependence helps or harms the decision.

\subsection{Distributed Reasoning Backrooms}

A multi-agent system is $\Sigma=(V,E,\sigma)$, where $V$ is the agent set, $E$ is the message topology, and $\sigma(w)$ is the skill set assigned to agent $w\in V$. Suppose a designated source agent $j\in V$ receives $s_v$. Let $d_v^{\Sigma}$ denote the team's final decision and let $d_{v,-(j,s)}^{\Sigma}$ denote the matched decision after removing only $s_v$ from $\sigma(j)$. Cross-agent reliance is
{\small
\begin{equation}
r_v^{\Sigma}(s,j,x)=
\mathbb{1}\!\left[d_v^{\Sigma}\neq d_{v,-(j,s)}^{\Sigma}\right].
\end{equation}
}
When $r_v^{\Sigma}=1$, skill influence has propagated from its source into the team decision. The corresponding causal provenance is $p_v=(s_v,j)$. The final attribution response instead exposes the claimed provenance $\hat p_v$. We define the instance-level laundering indicator as
{\small
\begin{equation}
\ell_v(s,j,x)=r_v^{\Sigma}(s,j,x)
\mathbb{1}\!\left[\hat p_v(s,x)\neq(s_v,j)\right].
\end{equation}
}
Skill laundering therefore describes a decision that remains sensitive to the source skill even though the team fails to recover its identity and origin. For any agent $w\in V$, let $d_{v,-w}^{\Sigma}$ be the matched team decision after removing $w$ while preserving the remaining organization. Agent-removal counterfactuals define the causal contribution profile
{\small
\begin{equation}
c_v(w,x)=\mathbb{1}\!\left[d_v^{\Sigma}\neq d_{v,-w}^{\Sigma}\right],
\qquad w\in V.
\end{equation}
}
Disagreement between this profile and the team's stated division of labor is \emph{contribution mismatch}. These definitions separate skill influence, effect direction, and provenance before benchmark aggregation.

\section{Methodology}

We propose \backtrace{}, a causal evaluation framework that tests and evaluates claimed skill use against intervention-measured influence. \bench{} instantiates it over tasks, frozen artifacts, intervention conditions, system organizations, and scoring code. Separating framework from benchmark enables the same audit across skills and domains.

\subsection{\backtrace{} Audit Protocol}

Given a frozen system, task distribution, and named skill, \backtrace{} asks whether changing that artifact changes the decision and whether subsequent attribution identifies the influence. It requires a stable answer normalizer or verifier but no access to model internals. Every comparison fixes the task instance, model, prompt frame, decoding policy, and organization while varying only the skill or its assignment.

\paragraph{Paired execution.}
For each instance, the audit records the no-skill and every skill-conditioned decision before asking about attribution. Once each answer is committed, a schema-constrained query elicits claimed skill use and, for teams, its source role; the no-skill run uses the same schema to expose unsupported provenance. We then join these records by instance to measure utility, answer sensitivity, attribution, and provenance.

\paragraph{Skill interventions.}
To isolate which aspect of a supplied skill drives a decision, we construct matched interventions that vary its presence, wording, displayed identity, and procedural content while holding the task and system fixed. Removing the artifact defines the shared no-skill counterfactual; six skill-bearing variants probe distinct artifact properties, as Table~\ref{tab:conditions} summarizes. Appendix~\ref{app:intervention-examples} gives a concrete example and its cross-domain counterpart.

\begin{table*}[t]
\centering
\footnotesize
\setlength{\tabcolsep}{4pt}
\begin{tabular}{@{}lp{0.3\textwidth}p{0.22\textwidth}p{0.32\textwidth}@{}}
\toprule
Condition & Intervention & Preserved control & Identified behavior \\
\midrule
None & delete the skill artifact & task, system, and organization & paired counterfactual baseline \\
Correct & supply the valid domain procedure & task and system & utility and answer reliance \\
Paraphrase & rewrite without changing meaning & procedural semantics & stability to surface wording \\
Misleading & permit reverse rule application & name and remaining scaffold & literal execution versus selective use \\
Name swap & relabel the correct procedure & procedural body & whether effects survive relabeling \\
Content swap & replace the body under the correct name & displayed identity & whether effects follow content \\
Irrelevant & supply a complete other-domain procedure & artifact structure & hidden effects and citation discipline \\
\bottomrule
\end{tabular}
\caption{\backtrace{} skill interventions and the artifact property isolated by each contrast.}
\label{tab:conditions}
\end{table*}

\paragraph{Validation and scoring.}
Answer and attribution parsers validate each record before scoring. We allow at most one format-only repair, which cannot re-solve the task. Unresolved final answers count as incorrect for accuracy; causal and attribution metrics exclude invalid pairs and report their denominators. All conditions for a model share the backend, decoding policy, context budget, and generation budget. Appendix~\ref{app:audit} gives the algorithm, design assumptions, recovery policy, and execution settings.

\paragraph{\bench{} construction.}
\bench{} instantiates the audit with 300 controlled PrOntoQA-style Logic problems \citep{saparov2023language}, 283 natural MATH-500 problems \citep{hendrycks2021measuring}, seven intervention conditions, and matched single- and multi-agent organizations. Logic balances five proof depths; reversing one designated rule leads into a checker-verified wrong-answer chain, so we can test whether agents follow misleading skill content. Mathematics tests the same gap without a constructed trigger. Frozen artifacts and identical instances are reused across models and organizations to avoid task-selection confounds. Appendix~\ref{app:audit} details generation, sampling, schemas, verification, and roles.

\begin{figure*}[htbp]
\centering
\includegraphics[width=\textwidth]{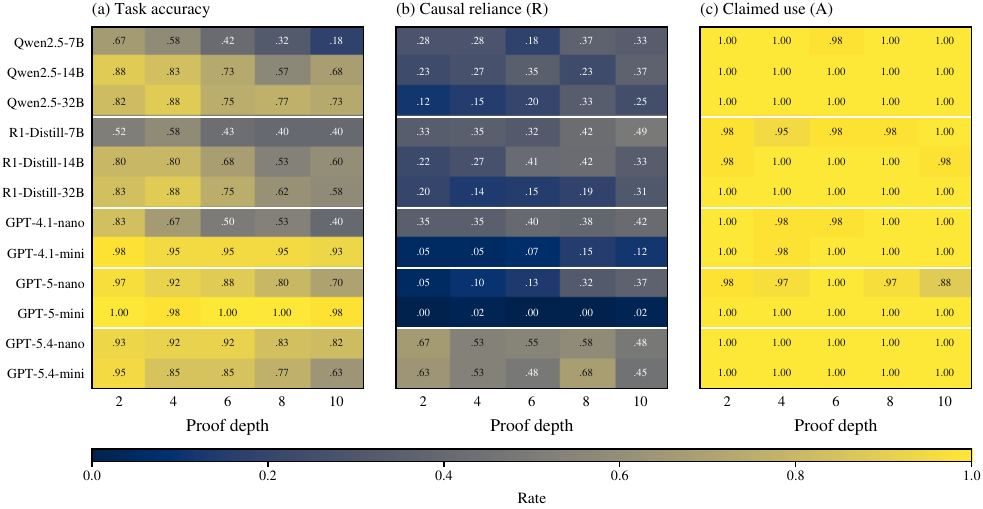}
\caption{Correct-skill results across Logic proof depths (60 instances per depth).}
\label{fig:depth}
\end{figure*}

\subsection{Evaluation Metrics}

Let $N_{\mathrm{all}}$ denote the full predeclared cohort. For a skill-bearing condition $v$, $\mathcal{I}_v$ contains instances with normalized skill and no-skill answers and a valid attribution record, and $N_v=\lvert\mathcal{I}_v\rvert$. Accuracy uses $N_{\mathrm{all}}$, while causal and attribution metrics use $N_v$; distributed metrics use their reported valid intersections. This aggregation separates whether the artifact changes the answer, whether that change helps, and whether stated skill use matches the dependence.

\paragraph{Reliance, attribution, and utility.}
The reliance rate and observable attribution rate are
{\small
\begin{equation}
R_v=\frac{1}{N_v}\sum_{i\in\mathcal{I}_v}r_v(s,x_i),
\qquad
A_v=\frac{1}{N_v}\sum_{i\in\mathcal{I}_v}a_v(s,x_i).
\end{equation}
}
Both are descriptive. High reliance can reflect helpful guidance or harmful interference, while high attribution can reflect faithful use or indiscriminate citation. Let $z_v(i)=\mathbb{1}[d_v(x_i)=d_i^*]$ when the answer parses and $z_v(i)=0$ otherwise. Accuracy is $\mathrm{Acc}_v=N_{\mathrm{all}}^{-1}\sum_{i=1}^{N_{\mathrm{all}}}z_v(i)$, where $d_i^*=\nu(y_i^*)$, and $\mathrm{Acc}_{\varnothing}$ is defined analogously. We present their signed difference alongside reliance and attribution,
{\small
\begin{equation}
\Delta\mathrm{Acc}_v
=\mathrm{Acc}_v-\mathrm{Acc}_{\varnothing}
=\frac{1}{N_{\mathrm{all}}}\sum_{i=1}^{N_{\mathrm{all}}}
\bigl(z_v(i)-z_{\varnothing}(i)\bigr).
\end{equation}
}

\paragraph{Attribution fidelity.}
For a fixed condition, let $n_{ra}$ count instances with reliance $r$ and attribution $a$. The Backroom Gap is the empirical disagreement rate
{\small
\begin{equation}
\label{def:gap}
\Gamma_v=\frac{n_{10}+n_{01}}{N_v}
=\frac{1}{N_v}\sum_{i\in\mathcal{I}_v}m_v(s,x_i).
\end{equation}
}
Lower $\Gamma$ is better. Its two directions are summarized by the silent-uptake rate and performative-use rate
{\small
\begin{equation}
\mathrm{SUR}=\frac{n_{10}}{n_{10}+n_{11}},
\qquad
\mathrm{PUR}=\frac{n_{01}}{n_{01}+n_{11}},
\end{equation}
}
when their denominators are nonzero. SUR and PUR are conditional error rates and do not add to $\Gamma$.

We summarize overlap between measured reliance and claimed use with Attribution Fidelity Score. For two nonnegative values, $\operatorname{HMean}(z_1,z_2)=2z_1z_2/(z_1+z_2)$ denotes their harmonic mean whenever the denominator is positive.
{\small
\begin{equation}
\label{def:afs}
\mathrm{AFS}=
\frac{2n_{11}}{2n_{11}+n_{10}+n_{01}}
=\operatorname{HMean}(1-\mathrm{SUR},1-\mathrm{PUR}).
\end{equation}
}
AFS ranges from 0 to 1 and is higher-is-better. It is undefined when neither reliance nor attribution occurs. Unlike an aggregate of performance and attribution, AFS does not reward a skill merely for changing many answers.

\paragraph{Distributed attribution.}
For the designated source agent $j$ and a valid distributed intersection of size $N$, we define cross-agent propagation, laundering, provenance recovery, and false provenance as
{\small
\begin{equation}
\begin{aligned}
\mathrm{CAP}_v&=\frac{1}{N}\sum_i r_v^{\Sigma}(s,j,x_i),\\
\mathrm{LR}_v&=\frac{1}{N}\sum_i \ell_v(s,j,x_i),\\
\mathrm{PR}_v&=\frac{1}{N}\sum_i
\mathbb{1}[\hat p_v(s,x_i)=(s_v,j)],\\
\mathrm{FPR}&=\frac{1}{N}\sum_i
\mathbb{1}[q_i^{\varnothing}\text{ names any skill or source}].
\end{aligned}
\end{equation}
}
Here LR is the joint rate at which the source skill changes the team decision but final attribution misses its skill--source pair. Hence $\mathrm{LR}_v\leq\mathrm{CAP}_v$; LR is not $1-\mathrm{PR}$ because PR also evaluates causally inactive instances. The variable $q_i^{\varnothing}$ denotes the no-skill attribution response, so FPR captures unsupported skill naming rather than false recovery of $s_v$ specifically. To compare organizations on the same skill and paired cohort, we use $\mathrm{OE}_v=\mathrm{CAP}_v-R_v$. Positive values denote amplification of answer sensitivity and negative values attenuation. We compare the intervention-based contribution vector $\mathbf{c}_v(x_i)=(c_v(w,x_i))_{w\in V}$ against the team's stated roles.

\section{Experiments}

\subsection{Experimental Setup}

\paragraph{Models and domains.}
We evaluate Qwen2.5-Instruct and DeepSeek-R1-Distill-Qwen at 7B, 14B, and 32B \citep{qwen2025qwen25,deepseek2025r1}, together with the nano and mini variants of GPT-4.1, GPT-5, and GPT-5.4. All twelve models run the 300-instance Logic audit under the shared no-skill baseline and six skill-bearing conditions. A additional six-model intersection runs both the complete 283-problem Mathematics audit for generalization analysis.

\paragraph{Paired evaluation.}
Each skill-conditioned answer is paired by instance with the same no-skill answer, normalized to a domain-specific decision. After the decision is committed, the agent identifies which supplied skill it used. The all-condition tables jointly present signed utility $\Delta\mathrm{Acc}$ and Attribution Fidelity Score (AFS). The former captures whether a skill helps or harms, while the latter measures instance-level alignment between intervention-defined reliance and claimed use. The depth analysis reports accuracy, causal reliance $R$, and claimed use $A$ separately. Accuracy retains every terminal generation, whereas reliance and attribution metrics use valid parsed pairs and report their effective denominators. All conditions within a model-domain comparison share the same prompt frame, backend, decoding policy, and budgets. Each condition covers its full predeclared cohort; conclusions rely on paired patterns repeated across models and conditions. Appendix~\ref{app:audit} gives execution details. Complete metric decompositions appear in Appendix~\ref{app:complete-results}; further controls are organized by RQ thereafter.

\subsection{RQ1: Do Agents Exhibit a Reasoning Backroom?}

\begin{table*}[htbp]
\centering
\footnotesize
\setlength{\tabcolsep}{1.7pt}
\begin{tabular}{@{}lc cc cc cc cc cc cc@{}}
\toprule
 & & \multicolumn{2}{c}{Correct} & \multicolumn{2}{c}{Paraphrase} & \multicolumn{2}{c}{Misleading} & \multicolumn{2}{c}{Name swap} & \multicolumn{2}{c}{Content swap} & \multicolumn{2}{c}{Irrelevant} \\
\cmidrule(lr){3-4} \cmidrule(lr){5-6} \cmidrule(lr){7-8} \cmidrule(lr){9-10} \cmidrule(lr){11-12} \cmidrule(l){13-14}
Model & Base & $\Delta$Acc & AFS$\uparrow$ & $\Delta$Acc & AFS$\uparrow$ & $\Delta$Acc & AFS$\uparrow$ & $\Delta$Acc & AFS$\uparrow$ & $\Delta$Acc & AFS$\uparrow$ & $\Delta$Acc & AFS$\uparrow$ \\
\midrule
\rowcolor{black!5} \textsc{Qwen}-7B  & .48 & -.05 & .45 & -.03 & .53 & -.01 & .50 & +.02 & .51 & -.01 & .47 & -.02 & .53 \\
\rowcolor{black!5} \textsc{Qwen}-14B & .57 & +.17 & .45 & +.14 & .49 & +.15 & .47 & +.16 & .45 & +.01 & .38 & +.01 & .00 \\
\rowcolor{black!5} \textsc{Qwen}-32B & .79 & .00 & .35 & +.03 & .33 & .00 & .36 & +.03 & .34 & -.01 & .30 & -.02 & .28 \\
\specialrule{0.5pt}{1.0pt}{1.0pt}
\rowcolor{black!9} \textsc{DSR1}-7B  & .47 & -.01 & .54 & +.02 & .53 & +.01 & .54 & -.03 & .45 & +.01 & .53 & +.05 & .44 \\
\rowcolor{black!9} \textsc{DSR1}-14B & .68 & .00 & .50 & +.04 & .48 & .00 & .51 & +.01 & .52 & -.04 & .37 & -.10 & .29 \\
\rowcolor{black!9} \textsc{DSR1}-32B & .71 & +.03 & .33 & +.05 & .31 & +.02 & .26 & +.04 & .26 & +.01 & .29 & -.03 & .31 \\
\specialrule{0.5pt}{1.0pt}{1.0pt}
\rowcolor{black!5} \textsc{GPT-4.1}-nano & .58 & .00 & .55 & +.02 & .54 & -.01 & .50 & +.01 & .48 & -.03 & .57 & -.07 & .52 \\
\rowcolor{black!5} \textsc{GPT-4.1}-mini & .94 & +.02 & .15 & +.02 & .14 & +.03 & .11 & +.02 & .15 & .00 & .13 & -.02 & .03 \\
\specialrule{0.5pt}{1.0pt}{1.0pt}
\rowcolor{black!9} \textsc{GPT-5}-nano & .86 & -.01 & .30 & -.05 & .34 & -.01 & .34 & -.01 & .32 & -.02 & .22 & +.01 & .20 \\
\rowcolor{black!9} \textsc{GPT-5}-mini & 1.00 & -.01 & .01 & -.01 & .02 & -.19 & .31 & -.01 & .02 & .00 & .01 & -.01 & .00 \\
\specialrule{0.5pt}{1.0pt}{1.0pt}
\rowcolor{black!5} \textsc{GPT-5.4}-nano & .39 & +.49 & .72 & +.47 & .69 & +.46 & .69 & +.43 & .65 & +.07 & .18 & +.02 & .17 \\
\rowcolor{black!5} \textsc{GPT-5.4}-mini & .29 & +.52 & .72 & +.56 & .74 & +.50 & .70 & +.50 & .69 & -.01 & .24 & .00 & .19 \\
\bottomrule
\end{tabular}
\caption{Logic single-agent results across all skill interventions (300 instances per condition).}
\label{tab:logic-all-conditions}
\end{table*}

\begin{table*}[htbp]
\centering
\footnotesize
\setlength{\tabcolsep}{1.7pt}
\begin{tabular}{@{}lc cc cc cc cc cc cc@{}}
\toprule
 & & \multicolumn{2}{c}{Correct} & \multicolumn{2}{c}{Paraphrase} & \multicolumn{2}{c}{Misleading} & \multicolumn{2}{c}{Name swap} & \multicolumn{2}{c}{Content swap} & \multicolumn{2}{c}{Irrelevant} \\
\cmidrule(lr){3-4} \cmidrule(lr){5-6} \cmidrule(lr){7-8} \cmidrule(lr){9-10} \cmidrule(lr){11-12} \cmidrule(l){13-14}
Model & Base & $\Delta$Acc & AFS$\uparrow$ & $\Delta$Acc & AFS$\uparrow$ & $\Delta$Acc & AFS$\uparrow$ & $\Delta$Acc & AFS$\uparrow$ & $\Delta$Acc & AFS$\uparrow$ & $\Delta$Acc & AFS$\uparrow$ \\
\midrule
\rowcolor{black!5} \textsc{Qwen}-14B & .78 & .00 & .26 & -.01 & .29 & -.02 & .32 & -.02 & .29 & .00 & .23 & -.01 & .00 \\
\rowcolor{black!5} \textsc{Qwen}-32B & .77 & -.02 & .27 & -.02 & .25 & +.02 & .29 & .00 & .29 & .00 & .22 & .00 & .00 \\
\specialrule{0.5pt}{1.0pt}{1.0pt}
\rowcolor{black!9} \textsc{DSR1}-14B & .84 & +.03 & .20 & +.03 & .18 & +.05 & .14 & +.02 & .17 & .00 & .18 & +.03 & .22 \\
\specialrule{0.5pt}{1.0pt}{1.0pt}
\rowcolor{black!5} \textsc{GPT-4.1}-mini & .86 & -.01 & .17 & .00 & .16 & .00 & .14 & -.01 & .11 & .00 & .11 & .00 & .00 \\
\specialrule{0.5pt}{1.0pt}{1.0pt}
\rowcolor{black!9} \textsc{GPT-5}-mini & .87 & .00 & .14 & -.02 & .11 & .00 & .11 & +.01 & .14 & +.01 & .11 & -.01 & .18 \\
\specialrule{0.5pt}{1.0pt}{1.0pt}
\rowcolor{black!5} \textsc{GPT-5.4}-nano & .78 & +.01 & .33 & +.04 & .27 & +.01 & .31 & +.05 & .26 & +.02 & .27 & +.02 & .43 \\
\bottomrule
\end{tabular}
\caption{Mathematics single-agent results across all skill interventions (283 instances per condition).}
\label{tab:math-all-conditions}
\end{table*}

\paragraph{Logic.}
\textbf{\emph{Skill-augmented agents exhibit a clear Reasoning Backroom.}} Figure~\ref{fig:depth} shows nearly saturated correct-skill attribution even as accuracy and removal-defined reliance vary across models and proof depths. This decomposition directly reveals stable claims despite changing causal dependence. Table~\ref{tab:logic-all-conditions} extends the comparison across all interventions with AFS. Its sharp variation shows that claimed use tracks reliance unreliably, even under the same valid procedure. Silent uptake under swapped and irrelevant artifacts further shows that the mismatch runs in both directions. It also persists under paraphrased and misleading artifacts, so it is not specific to identity or content swaps. Crucially, joint elicitation leaves attribution at $.98$--$1.00$; even at perfect baseline accuracy, eight models claim use despite zero deletion reliance. Thus neither query order nor task failure explains the observed saturation (Appendix~\ref{app:rq1-rq4}).

\paragraph{Generalization to Mathematics.}
\textbf{\emph{The Reasoning Backroom generalizes from controlled Logic to natural Mathematics.}} Table~\ref{tab:math-all-conditions} shows the same separation without a constructed trigger. No model--condition pair exceeds $.43$ AFS, and low fidelity spans valid, paraphrased, altered, and irrelevant artifacts. The result is therefore neither created by an adversarial condition nor explained by average performance shifts. What transfers is a direct failure of claimed use to distinguish decisions that change with the skill from those that do not. Together, the two domains establish the Backroom as a recurring property of skill attribution across controlled and natural tasks, rather than a quirk of one benchmark construction.

\subsection{RQ2: What Determines Skill Influence?}

\textbf{\emph{Procedural content, rather than wording or displayed identity, carries most skill influence.}} In Table~\ref{tab:logic-all-conditions}, paraphrasing the procedure or changing its name largely preserves the direction and magnitude of the correct-skill effect across model rows. Replacing the body under the original name breaks that correspondence and collapses the largest gains. Comparing Name swap, which preserves content, with Content swap, which preserves identity, shows that the procedural body drives the behavioral effect. AFS remains low under the identity controls, showing that claimed use does not identify which artifact component changed the decision.

\begin{figure*}[t]
\centering
\includegraphics[width=0.98\textwidth]{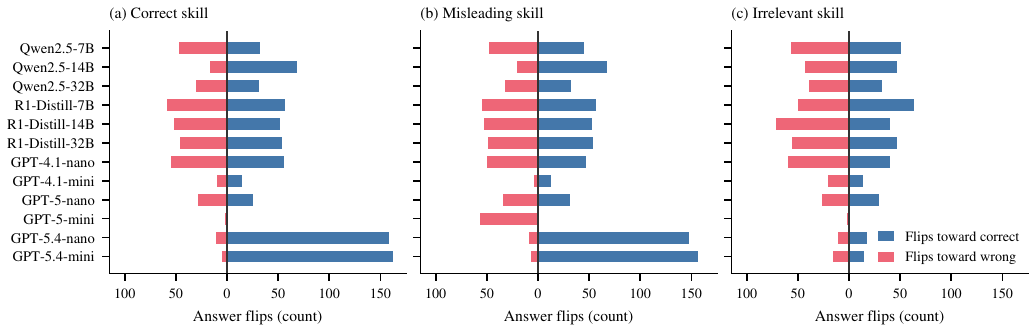}
\caption{Logic answer-flip directions relative to paired no-skill runs; right bars improve correctness and left bars reduce it.}
\label{fig:flips}
\end{figure*}

\textbf{\emph{Model-specific execution determines whether that content helps, harms, or cancels out.}} Misleading preserves the scaffold but permits reverse rule application into a verified wrong-answer path. Most models nevertheless retain a profile close to the correct procedure, whereas GPT-5-mini converts the same intervention into systematic harm. Figure~\ref{fig:flips} explains this divergence: comparable amounts of answer change can be dominated by helpful flips, harmful flips, or opposing flips that cancel in aggregate. Agents therefore do not apply a skill as an all-or-nothing capability. This pattern is consistent with selective application or over-extension. Procedural content constrains available influence, while model-specific execution determines its direction. Because $\Delta\mathrm{Acc}$ is the signed balance of helpful and harmful flips, near-zero utility can conceal substantial decision churn.

\subsection{RQ3: Can MAS Preserve Skill Provenance?}

\textbf{\emph{Skill influence can propagate while its causal provenance is lost.}} In Table~\ref{tab:mas-crossdomain}, LR is often nearly as large as CAP across both domains. Because LR counts propagation events whose exact skill--source pair is not recovered, the gap between CAP and LR is the propagated influence whose provenance survives. For most systems this gap is negligible: the source skill changes the team decision, yet the team fails to identify it afterward. Conversely, GPT-5-mini frequently names a source on Logic although source-skill removal rarely changes the decision. When influence persists but the named skill or source is wrong, this is skill laundering.

\begin{table}[htbp]
\centering
\footnotesize
\setlength{\tabcolsep}{1.5pt}
\begin{tabular}{@{}l cccc cccc@{}}
\toprule
& \multicolumn{4}{c}{Logic} & \multicolumn{4}{c}{Mathematics} \\
\cmidrule(lr){2-5} \cmidrule(l){6-9}
Model & CAP & LR & PR & FPR & CAP & LR & PR & FPR \\
\midrule
\rowcolor{black!5} \textsc{Qwen}-14B & .33 & .33 & .01 & 1.00 & .13 & .13 & .00 & 1.00 \\
\rowcolor{black!5} \textsc{Qwen}-32B & .25 & .20 & .17 & 1.00 & .17 & .16 & .01 & 1.00 \\
\specialrule{0.5pt}{1.0pt}{1.0pt}
\rowcolor{black!9} \textsc{DSR1}-14B & .39 & .39 & .00 & .97 & .15 & .15 & .00 & .77 \\
\specialrule{0.5pt}{1.0pt}{1.0pt}
\rowcolor{black!5} \textsc{GPT-4.1}-mini & .03 & .03 & .08 & 1.00 & .11 & .09 & .20 & 1.00 \\
\specialrule{0.5pt}{1.0pt}{1.0pt}
\rowcolor{black!9} \textsc{GPT-5}-mini & .01 & .00 & .71 & 1.00 & .09 & .09 & .01 & .99 \\
\specialrule{0.5pt}{1.0pt}{1.0pt}
\rowcolor{black!5} \textsc{GPT-5.4}-nano & .11 & .11 & .07 & 1.00 & .17 & .17 & .00 & .99 \\
\bottomrule
\end{tabular}
\caption{Multi-agent results on 150 Logic and 142 Mathematics instances; LR is propagated influence without exact skill--source recovery.}
\label{tab:mas-crossdomain}
\end{table}

\begin{figure}[htbp]
\centering
\includegraphics[width=\columnwidth]{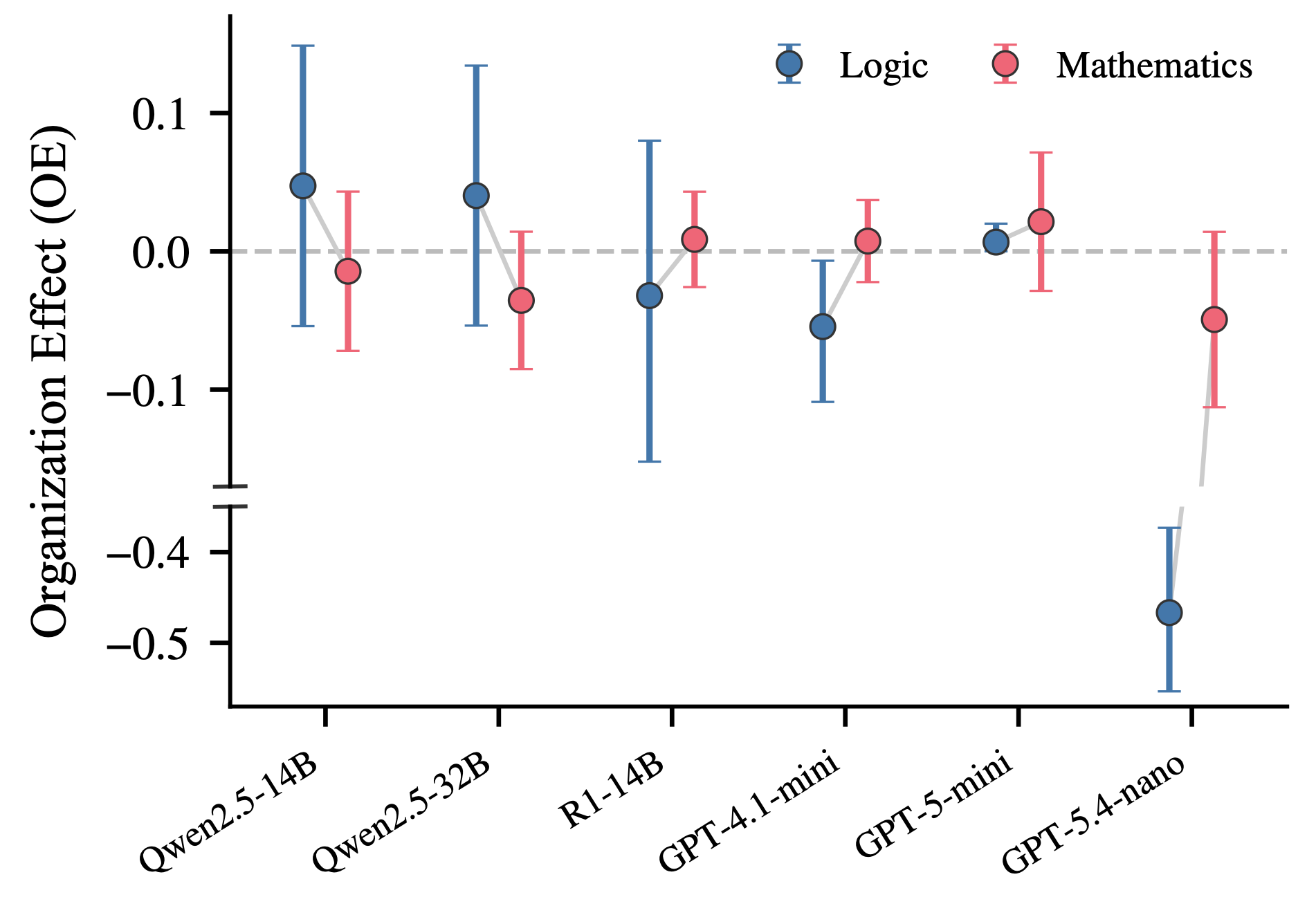}
\caption{Organization effects for the six shared systems with a broken y-axis. Error bars show 95\% paired-bootstrap intervals; lines connect domains within each model.}
\label{fig:mas-organization}
\end{figure}

\textbf{\emph{A plausible source attribution is not causal evidence.}} Skill naming remains pervasive in no-skill teams throughout Table~\ref{tab:mas-crossdomain}, showing that a coherent provenance narrative can arise without any supplied artifact. High FPR directly measures unsupported provenance construction when no skill or source exists. Together, LR close to CAP and pervasive FPR expose a two-sided failure. Teams lose provenance when influence is present, yet construct it when no artifact was supplied. Figure~\ref{fig:mas-organization} shows that team organization can amplify or attenuate skill sensitivity, with values near zero preserving the single-agent effect. Distribution attenuates some Logic effects, mildly amplifies others, and yields no clear Mathematics shift. Thus teams do not consistently amplify single-agent effects. Provenance is lost whether organization amplifies or attenuates the effect. The distributed Backroom is therefore an AI-provenance failure, not merely a communication effect.

\subsection{RQ4: Can Observation Replace Intervention?}

\textbf{\emph{The tested visible signals do not reliably identify whether a skill changed an agent's decision.}} Table~\ref{tab:detectors} compares four observational signals against deletion-defined reliance. Although they flag runs at very different rates, their precision remains near the $.34$ prevalence of reliance. At this base rate, a flagged run is no more likely than an average run to show actual reliance. Explicit attribution achieves high recall by labeling almost every run, while direct mentions flag few runs and miss most reliance events. Semantic similarity and the LLM judge fall between these extremes without better separating reliance from non-reliance. Thus prediction frequency changes across signals, but the ability to identify reliance does not, because a single execution lacks the counterfactual decision.

\begin{table}[htbp]
\centering
\footnotesize
\setlength{\tabcolsep}{4.2pt}
\begin{tabular}{@{}lcccc@{}}
\toprule
Observable signal & Prec. & Recall & F1 & Pos. rate \\
\midrule
Explicit attribution & .32 & .90 & .47 & .95 \\
Skill mention & .37 & .08 & .13 & .07 \\
Trace--skill similarity & .31 & .45 & .37 & .49 \\
LLM judge & .31 & .69 & .43 & .76 \\
\bottomrule
\end{tabular}
\caption{Observational detectors of deletion-defined skill reliance on 3,600 Logic runs.}
\label{tab:detectors}
\end{table}

\textbf{\emph{Performance does not reveal whether a skill changed the decision.}} Outcome improvement already requires a paired no-skill result and records only the net beneficial effect. It misses harmful and correctness-preserving answer changes and can vanish when helpful and harmful flips cancel, as Figure~\ref{fig:flips} demonstrates. The detectors in Table~\ref{tab:detectors} observe only one execution and cannot determine whether a skill changed a particular decision. \textbf{\emph{Confirming dependence requires intervention.}}

\section{Conclusion}

We evaluate whether skill-augmented agents expose the causal influence of external procedures in single- and multi-agent systems. We find a pervasive Reasoning Backroom: attribution diverges from decision dependence, effects follow procedural content rather than displayed identity, and propagated influence loses provenance. Our interventions test whether the claimed skill and source actually changed the decision. Our answer-level audit cannot recover internal mechanisms or exclude equivalent pretrained procedures, and covers only frozen textual skills, two domains, selected systems, and mostly deterministic executions. Future work should audit adaptive, long-lived agents.

\bibliography{aaai2027}

\clearpage
\appendix
\setcounter{secnumdepth}{1}

\section{Detailed Audit Protocol}
\label{app:audit}

The \backtrace{} audit operationalizes a causal comparison without treating a model's visible rationale as ground truth. It stores decisions and attribution responses as separate records, validates both channels, and computes attribution only after every skill condition has been paired with the same no-skill counterfactual.

In Algorithm~\ref{alg:audit}, $i$ indexes instances and $v$ indexes intervention conditions. For any per-instance quantity $z_v(x_i)$, the shorthand $z_i^v$ places the instance in the subscript and the condition in the superscript. The fixed execution context $\omega$ is omitted. The flag $f_i^{\varnothing}$ equals one when the no-skill attribution response for instance $i$ names any skill or source.

\subsection{\backtrace{} Design Commitments}
\label{app:design-commitments}

Four commitments bound the causal interpretation. \emph{Identifiability} holds the instance, model, prompt frame, decoding policy, and organization fixed while changing only the audited artifact or its assignment. \emph{Decision verifiability} compares normalized task decisions rather than lexical differences between free-form outputs. \emph{Semantic control} uses meaning-preserving and identity-swapping variants to distinguish procedural content from wording and labels. \emph{Attribution hygiene} commits the decision before eliciting claimed skill use, preventing the attribution query from changing the answer it describes. The design can extend to tool permissions, parameters, composition, or memory only when every new contrast has a predeclared interpretation and preserves all non-target inputs.

\subsection{\bench{} Construction Details}
\label{app:benchmark-construction}

\paragraph{Controlled logic track.}
We generate 300 first-order taxonomy problems in the style of PrOntoQA \citep{saparov2023language} with seed 20260715. The set is balanced across proof depths $\{2,4,6,8,10\}$ with 60 instances at each depth and balanced truth labels. Fictional vocabulary limits recall shortcuts, every label is provable by construction, and every instance stores a gold proof. Three distractor chains terminate in the opposite polarity and connect to the main chain. Forward-only application remains valid, while reversing a designated cross-link enters a constructed wrong-answer chain. A symbolic checker verifies the label, proof, and corruption target for every instance.

\paragraph{Natural mathematics track.}
We select 283 competition problems from MATH-500 \citep{hendrycks2021measuring}, stratified over difficulty levels 1 through 5. The subset is fixed before inference by shuffling within each level with seed 20260715 and taking up to 60 problems per level. Final expressions are normalized before pairing. Unlike the synthetic logic track, these human-authored problems do not guarantee that a supplied procedure is needed on every instance. They test whether the attribution gap persists when skills interact with heterogeneous natural problems rather than a known construction.

\paragraph{Frozen skill library.}
Each artifact contains a stable display name, purpose, applicability conditions, procedure, prohibited shortcuts, and expected output format. It never contains an instance answer. The logic library covers forward chaining, contradiction checking, and invalid-inference avoidance. The mathematics library provides problem decomposition and independent verification. The irrelevant condition supplies a complete but task-inapplicable skill from the frozen library rather than a nonsensical prompt. Every artifact is frozen before evaluation and reused across all models and organizations.

\subsection{Concrete Intervention Example}
\label{app:intervention-examples}

Table~\ref{tab:intervention-example} illustrates the complete Logic intervention family using the frozen Forward Chaining artifact. The displayed name is only the visible skill header, while the procedure supplied beneath it may differ. Contradiction Checking is not an intervention category, but a sibling verification skill in the frozen Logic library. Given a candidate answer, it attempts to derive the candidate's negation from the same facts and forward rules, revising the answer when the negation is derivable and otherwise retaining it. The two swap conditions deliberately separate visible identity from procedural content. Name swap preserves the Forward Chaining procedure under the Contradiction Checking header, whereas Content swap preserves the Forward Chaining header but supplies the Contradiction Checking procedure. Their comparison tests whether behavioral influence follows the supplied procedure rather than the displayed identity, and whether claimed use tracks that change. The remaining conditions isolate skill absence, meaning-preserving wording, a corrupted procedural constraint, and a complete but inapplicable procedure. Mathematics applies the same contrasts to Problem Decomposition.

\begin{table*}[t]
\centering
\footnotesize
\setlength{\tabcolsep}{4.5pt}
\begin{tabular}{@{}lp{0.20\textwidth}p{0.56\textwidth}@{}}
\toprule
Condition & Displayed name & Supplied procedure and intervention \\
\midrule
None & -- & provide no skill block while preserving the task and execution context \\
Correct & Forward Chaining & supply the original one-way forward-chaining procedure \\
Paraphrase & Forward Chaining & supply a meaning-preserving rewrite of the same procedure \\
Misleading & Forward Chaining & retain the scaffold but add a corrupted rule permitting reverse application from B to A \\
Name swap & Contradiction Checking & supply the Forward Chaining procedure but display Contradiction Checking \\
Content swap & Forward Chaining & supply the Contradiction Checking procedure but display Forward Chaining \\
Irrelevant & Problem Formalization & supply a complete but inapplicable procedure from the frozen library \\
\bottomrule
\end{tabular}
\caption{Concrete realization of the Logic intervention family.}
\label{tab:intervention-example}
\end{table*}

\begin{algorithm}[H]
\caption{The \backtrace{} paired audit protocol}
\label{alg:audit}
\textbf{Input} Frozen system $\pi$, pairs $\{(x_i,y_i^*)\}_{i=1}^{N}$, conditions $\mathcal{V}$\\
\textbf{Output} Paired decisions, attributions, provenance, and aggregate metrics
\begin{algorithmic}[1]
\FOR{$i=1$ \TO $N$}
    \STATE $d_i^*\leftarrow\nu(y_i^*)$
    \STATE $(\tau_i^{\varnothing},\hat y_i^{\varnothing})\leftarrow\pi_{\mathrm{ans}}(x_i,\varnothing)$
    \STATE $d_i^{\varnothing}\leftarrow\nu(\hat y_i^{\varnothing})$
    \STATE Commit $d_i^{\varnothing}$, then elicit and validate $q_i^{\varnothing}$
    \STATE $f_i^{\varnothing}\leftarrow\mathbb{1}[q_i^{\varnothing}\text{ names any provenance}]$
    \FOR{$v\in\mathcal{V}\setminus\{\varnothing\}$}
        \STATE $(\tau_i^v,\hat y_i^v)\leftarrow\pi_{\mathrm{ans}}(x_i,s_v)$
        \STATE $d_i^v\leftarrow\nu(\hat y_i^v)$
        \STATE Commit $d_i^v$, then elicit and validate $q_i^v$
        \STATE $(a_i^v,\hat p_i^v)\leftarrow g(q_i^v)$
        \STATE $r_i^v\leftarrow\mathbb{1}[d_i^v\neq d_i^{\varnothing}]$
        \STATE $u_i^v\leftarrow\mathbb{1}[d_i^v=d_i^*]-\mathbb{1}[d_i^{\varnothing}=d_i^*]$
    \ENDFOR
\ENDFOR
\STATE Join paired records and count the four $(r,a)$ states
\RETURN Utility, reliance, attribution, fidelity, and provenance metrics
\end{algorithmic}
\end{algorithm}

\paragraph{Organizations.}
Single-agent and multi-agent tracks reuse the same instances and skill variants. The multi-agent track deterministically takes every other instance from each frozen ordered cohort, yielding 150 Logic and 142 Mathematics instances; the same identifiers are reused across systems and paired single-agent comparisons. Logic teams assign fact, rule, verifier, and aggregator roles. Mathematics teams use solver, verifier, and aggregator roles. The primary distributed treatment assigns the audited skill only to its designated source while all agents exchange ordinary natural-language messages. Source-skill removal measures propagation, message interventions test whether influence survives transmission, and agent-removal counterfactuals recover a causal contribution profile. This paired construction prevents task selection from being confounded with system organization.

\paragraph{Execution environment.}
Local checkpoints are served with vLLM in bfloat16 on three NVIDIA A100-PCIE GPUs with 40\,GB memory each. The 32B checkpoints use tensor parallelism over two GPUs. Local inference uses greedy decoding. API requests set temperature to zero when the endpoint accepts that parameter and otherwise retain the provider-required default. The original runs use an 8192-token context window, while bounded corrective runs for truncated long reasoning use up to 24,576 tokens. Logic allocates 1024 output tokens to standard local models and up to 6144 to long-reasoning corrective runs. Mathematics allocates 2048 tokens to Qwen models and up to 8192 to long-reasoning DSR1 and GPT variants. Every condition within a model-domain audit uses the same backend and budget. Exact model revisions, commands, budgets, and recovery events are retained in the run manifests.

\paragraph{Uncertainty details.}
The presented rates use one deterministic decoding path per condition. Paired-bootstrap intervals elsewhere in the evaluation use 10,000 instance resamples and quantify instance uncertainty rather than serving variance.

\paragraph{Paired execution.}
For each instance, the None decision is generated once and reused as the baseline for every supplied artifact. This shared baseline makes each contrast differ only in the audited skill condition and prevents different comparison sets from producing artificial differences between variants. Instance identifiers, the model revision, prompt frame, organization, decoding settings, context budget, and output budget are frozen before inference.

\paragraph{Post-decision attribution.}
The attribution question is absent from the answer turn. The system first commits its reasoning trace and final decision, after which a separate request asks the system to identify the skills it used and their provenance. This ordering prevents the attribution request from changing the decision it is intended to describe. The None condition uses the same downstream attribution schema, so any named skill or source becomes a directly observable false-provenance event.

\paragraph{Validation and recovery.}
The domain parser maps each answer to a verifier representation, and the attribution parser maps each claimed name to the artifact list supplied in that run. Case and formatting are normalized without changing semantic content. Malformed telemetry receives at most a bounded format-only repair request that forbids re-solving. If validation still fails, the terminal generation remains in the condition file. Accuracy counts an unresolved answer as incorrect, whereas reliance, attribution, and four-state metrics exclude pairs lacking normalized answers or valid attribution records. Scoring still requires the None record and every required supplied-condition instance identifier; summaries expose answer and attribution parse rates together with each effective denominator.

\paragraph{Effective sample coverage.}
The final corrective Logic runs contain all 300 predeclared instance identifiers per condition. Answer parse rates span $.99$--$1.00$ for DSR1-7B, $.88$--$.97$ for DSR1-14B, and $.78$--$.87$ for DSR1-32B; attribution-record parse rates span $.99$--$1.00$, $.77$--$.97$, and $.76$--$.86$, respectively. Both GPT-5.4 variants attain a $1.00$ Logic answer-parse rate in the no-skill baseline and all six skill conditions, so their low baseline accuracy reflects valid normalized decisions rather than formatting failure. A truth-independent marginal bias alone would score $.50$ on the balanced labels. Instead, GPT-5.4-nano answers False on 113 of 150 gold-False and 145 of 150 gold-True instances; GPT-5.4-mini does so on 59 and 123, respectively. Their $.39$ and $.29$ accuracies thus reflect label-conditioned error asymmetry among valid outputs, not label skew alone. This pattern is consistent with the benchmark's verified opposite-polarity distractor paths, but does not by itself identify which path a model followed. The large gains under Correct, Paraphrase, and Name swap restore correct decisions rather than recover unparsed answers; Content swap and Irrelevant do not yield comparable gains, keeping the contrast content-sensitive. The DSR1-14B Mathematics rerun contains all 283 identifiers and reaches $.95$--$.99$ answer parsing and $.85$--$.97$ attribution-record parsing. Consequently, task accuracy retains the complete predeclared cohort, while causal and attribution statistics expose the valid paired cohort rather than silently treating truncated generations as parsed observations.

\paragraph{Multi-agent adaptation.}
For a team, $\pi$ in Algorithm~\ref{alg:audit} denotes the entire frozen organization rather than one model call. The ordinary run assigns the skill only to its designated source. The paired source-removal run deletes that artifact from the source while preserving the role graph and messages. The final attribution response must name both the skill and its source to count as provenance recovery. Agent-removal and message interventions are executed as additional paired conditions, leaving the scoring loop unchanged.

\paragraph{Interpretation boundary.}
The algorithm identifies answer sensitivity to a controlled artifact. It does not claim to reconstruct private token-level computation or that every possible attribution query must behave identically. Our conclusions concern the fixed post-decision attribution channel tested here. Residual backend nondeterminism can introduce answer variation even under fixed settings, so the released records retain model versions, decoding configurations, and all paired outputs needed to audit this assumption.

\subsection{Worked Audit Examples}
\label{app:worked-examples}

The following examples come from the released per-instance records. We preserve instance identifiers, normalized decisions, and structured attribution fields; task contexts and traces are abridged only by removing unrelated rules and repetition.

\paragraph{Silent uptake.}
For \textsc{Qwen}-14B on \texttt{logicv2-d2-t-017}, the verified proof is Alex $\rightarrow$ numpus $\rightarrow$ lorpusoid $\rightarrow$ shumpusling $\rightarrow$ loud, so the gold answer is True.
\begin{quote}
\small
\emph{No skill:} \texttt{Answer: False}.\\
\emph{Problem Formalization supplied:} \texttt{Answer: True}.\\
\end{quote}
The post-decision attribution lists no used skill and places Problem Formalization in the unused set. The artifact nevertheless changes the normalized decision and corrects the answer. Thus $r_v=1$, $a_v=0$, and $u_v=+1$, which is a silent-uptake event. The claim is answer-level sensitivity to the supplied artifact, not that the agent internally executed a quantitative formalization procedure.

\paragraph{Skill laundering.}
For the \textsc{Qwen}-14B team on \texttt{logicv2-d4-t-012}, only the rule specialist receives Forward Chaining. The verified path is Fae $\rightarrow$ rompusling $\rightarrow$ tumpusoid $\rightarrow$ shumpusoid $\rightarrow$ lempusette $\rightarrow$ grimpusoid $\rightarrow$ rough.
\begin{quote}
\small
\emph{Source-skill removal:} the rule specialist adds an unsupported rompusling $\rightarrow$ rompusoid step; the team answers False.\\
\emph{Forward Chaining at the rule specialist:} the specialist follows the verified path; the team answers True.\\
\emph{Final attribution:} most influential agent $=\texttt{rule}$; named skills $=\{\texttt{logical reasoning},\texttt{derivation}\}$; skill source $=\texttt{rule}$.
\end{quote}
Removing the source skill flips the team decision, so $r_v^{\Sigma}=1$. The final response retains the source role but replaces the exact artifact identity with generic descriptions, hence $\hat p_v\neq(\texttt{Forward Chaining},\texttt{rule})$ and $\ell_v=1$. This instance shows why naming a plausible contributor is weaker than recovering joint skill provenance.

\section{Relation to Chain-of-Thought Faithfulness}
\label{app:cot-faithfulness}

CoT faithfulness and skill provenance share an interventionist motivation but test different claims. CoT work asks whether a model-generated, instance-specific rationale reflects the factors or computation that drove its prediction \citep{turpin2023language,paul2024making,arcuschin2025chainofthought,hao-etal-2026-reasoning}. Our target is the relationship between an externally supplied, reusable artifact and the resulting decision. We ask whether changing that artifact changes the answer and whether the system preserves its identity and source when attributing the decision. The distinction is between explanation faithfulness and execution provenance. The first treats visible reasoning as evidence about a prediction process. The second treats a named capability artifact as an auditable input whose causal influence and provenance can be tested even when the visible trace is incomplete or misleading. Table~\ref{tab:cot-positioning} summarizes these differences.

\begin{table*}[t]
\centering
\small
\setlength{\tabcolsep}{5.0pt}
\begin{tabular}{@{}p{0.17\textwidth}p{0.36\textwidth}p{0.39\textwidth}@{}}
\toprule
Aspect & CoT faithfulness & Skill provenance in this work \\
\midrule
Claim under test & Whether a visible rationale reflects what drove one prediction & Whether a supplied skill changed a decision and its stated identity and source match that influence \\
Audited object & A generated, instance-bound trace or an input cue & A persistent external artifact and, in a team, its assignment \\
Counterfactuals & Perturb cues or generated reasoning and observe the prediction & Delete, paraphrase, corrupt, rename, substitute, or reassign the skill while holding the task and system fixed \\
Population target & Per-instance rationale faithfulness, then aggregated & An artifact-level influence profile over matched instances, including beneficial and harmful effects \\
Multi-agent target & Whether visible debate or messages preserve influential information & Propagation, loss of the skill and source association, and mismatch between stated and intervention-measured contributors \\
Interpretation & Evidence about explanation or trace validity & Evidence about capability provisioning, attribution, and execution provenance \\
\bottomrule
\end{tabular}
\caption{CoT faithfulness and skill provenance test complementary causal claims.}
\label{tab:cot-positioning}
\end{table*}

\paragraph{Why the two failure modes are independent.}
A faithful CoT can still support false skill provenance. The trace may accurately expose the derivation while the agent credits a named skill whose removal leaves the answer unchanged. Conversely, a trace may omit intermediate reasoning even when skill deletion flips the answer and the agent correctly names the artifact and its source. Thus neither axis entails the other. Our four attribution states test agreement between artifact-level decision dependence and stated skill use; they do not relabel trace quality as skill faithfulness.

\paragraph{What artifact-level intervention enables.}
A skill exists independently of the trace and recurs across instances. This permits matched deletion to measure answer sensitivity, helpful and harmful flips to separate dependence from utility, and crossed name and content interventions to identify whether behavior follows procedural semantics or displayed identity. Stable artifacts can also be moved between agents. Source removal, reassignment, and agent removal therefore test whether influence propagates, whether its origin is preserved, and whether attributed contributors match causal contributors. These questions have no required answer in a trace-only audit because a fluent rationale can reproduce a procedure without establishing which provisioned artifact supplied it.

\paragraph{Relation to multi-agent deception.}
Work on unfaithful reasoning and multi-agent belief manipulation demonstrates that visible text can conceal influential cues or strategically assemble truthful statements into a misleading account \citep{turpin2023language,hu-etal-2026-lying}. Our distributed result does not require strategic collusion, deceptive intent, or a false message. Skill laundering is defined behaviorally in ordinary task execution. Source-skill removal changes the team decision while the final attribution misses their joint identity. Skill-based provenance in no-skill teams further separates a plausible source narrative from actual provisioning. The contribution is therefore an audit of execution provenance, not a detector of deception.

\paragraph{Complementary scope.}
The artifact-level design provides a directly manipulable intervention target, not access to private reasoning. An unchanged answer cannot exclude an equivalent procedure learned during pretraining, and a changed answer identifies sensitivity under the fixed context rather than a token-level mechanism. CoT audits remain necessary when the scientific question concerns the truthfulness or completeness of the rationale itself. Used together, the two evaluations can distinguish whether a system exposes its reasoning process from whether it correctly accounts for an external capability's contribution.

\section{Additional Analysis for RQ2}
\label{app:rq2-controls}

\begin{figure*}[t]
\centering
\includegraphics[width=0.98\textwidth]{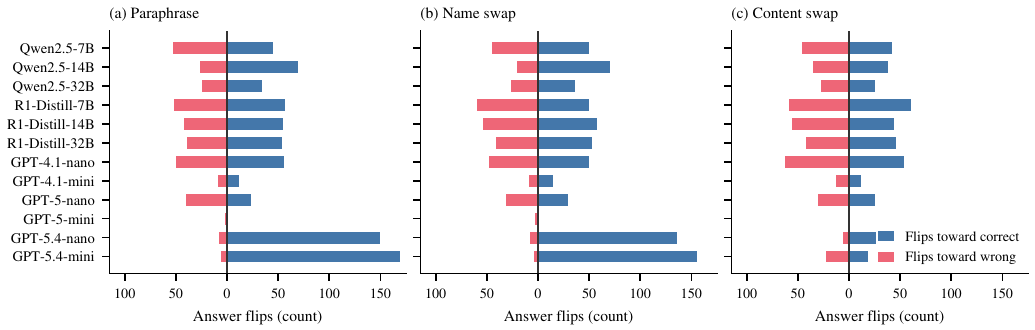}
\caption{Logic answer-flip directions for the remaining controls; right bars improve correctness and left bars reduce it.}
\label{fig:flip-controls}
\end{figure*}

\paragraph{Meaning-preserving and identity controls.}
Figure~\ref{fig:flip-controls} completes the six-condition directional audit. Paraphrase largely preserves the beneficial profile of the correct artifact: GPT-5.4-nano and GPT-5.4-mini produce 150/8 and 169/6 helpful/harmful flips, respectively. Name swap remains similarly asymmetric at 136/8 and 155/4, showing that these gains are not tied to the displayed name. Content swap sharply weakens the same pattern: the corresponding counts fall to 26/6 and 19/22. Under an equal-direction null, exact two-sided binomial tests reject symmetry for Paraphrase and Name swap in both models (all $p<3.6{\times}10^{-31}$) and for GPT-5.4-nano Content swap ($p=5.4{\times}10^{-4}$), but not GPT-5.4-mini Content swap ($p=.755$). These tests formalize flip directionality rather than serving variance. The crossed intervention identifies the procedural body, rather than wording or identity alone, as the component carrying the large behavioral effect.

\paragraph{Cancellation hidden by aggregate utility.}
Near-zero $\Delta\mathrm{Acc}$ does not imply that an artifact was behaviorally inert. Under Content swap, Qwen2.5-7B produces 42 helpful and 46 harmful flips, DSR1-7B produces 56 and 67, and GPT-4.1-nano produces 54 and 63. The opposing changes nearly cancel in aggregate even though many individual decisions differ from the no-skill baseline. Reading Figure~\ref{fig:flip-controls} with Table~\ref{tab:logic-all-conditions} therefore separates three quantities that cannot substitute for one another: whether the artifact changes an answer, whether the change helps, and whether the agent attributes the decision to that artifact.

\begin{figure*}[p]
\centering
\includegraphics[width=\textwidth]{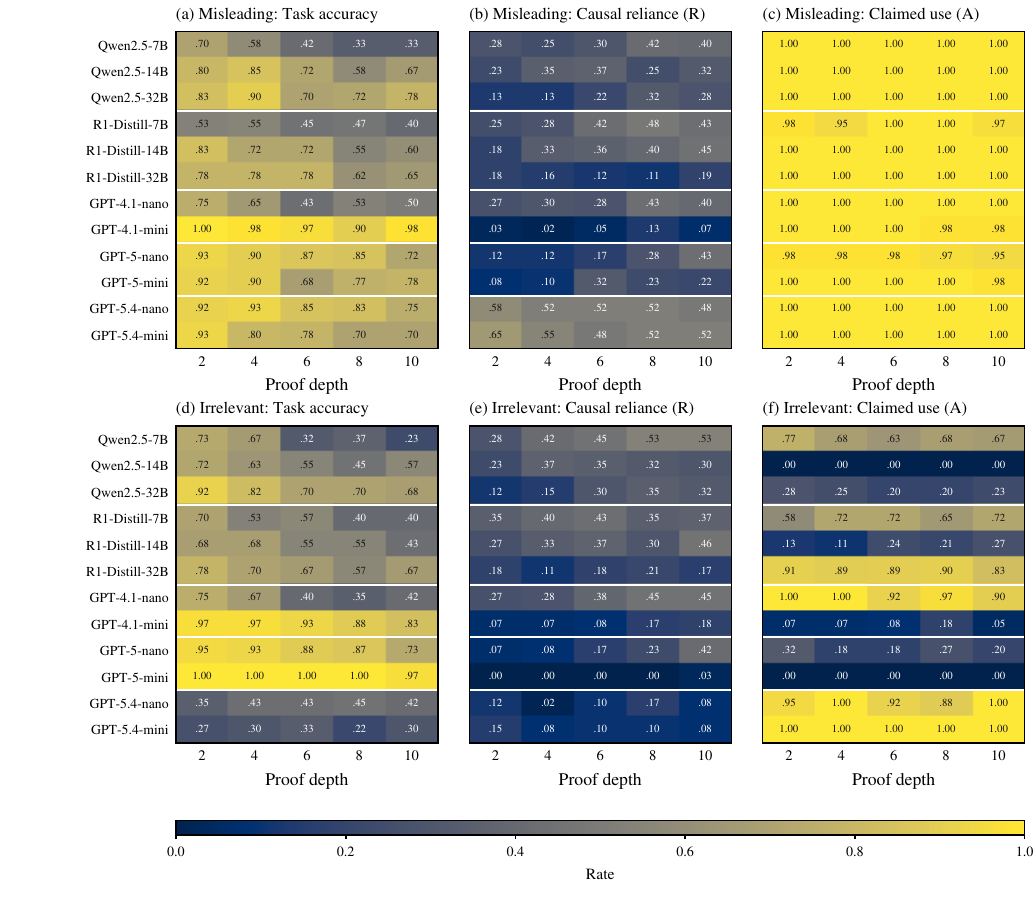}
\caption{Depth-resolved Logic results under misleading and irrelevant skills.}
\label{fig:depth-stress}
\end{figure*}

\paragraph{Depth robustness under stress skills.}
Figure~\ref{fig:depth-stress} tests whether the Backroom is confined to one difficulty level. Under the misleading skill, eleven models vary in $R$ by at least $.10$ across proof depths, with a median range of $.17$, while every model keeps $A$ within $.05$, with a median range of zero. The corrupted procedure therefore becomes more or less behaviorally active as reasoning depth changes without a matching shift in attribution. The irrelevant condition shows both directions of failure. Qwen-14B and GPT-5-mini attribute no use at any depth despite changing reliance, whereas GPT-5.4-mini attributes use throughout although $R$ remains only $.08$--$.15$. The mismatch is therefore stable across predeclared depth strata rather than driven by a single task regime.

\section{Additional Analysis for RQ3}
\label{app:rq3}

This section expands the cross-domain result in Table~\ref{tab:mas-crossdomain}. It reports the complete Logic cohort and then separates organization effects and role contribution. These controls explain how the distributed Backroom manifests without changing the main causal claim.

\paragraph{Contribution mismatch.}
Let $\mathcal{K}_v$ be the valid agent-removal cohort, $\mathcal{J}_v$ the skill runs with a valid reported most-influential specialist $\hat w_{v,i}$, and $\mathcal{W}=\{\mathrm{fact},\mathrm{rule},\mathrm{verifier}\}$ the removable Logic roles. We aggregate the instance-level contribution profile as
{\small
\begin{equation}
\begin{aligned}
\bar c_v(w)&=\frac{1}{|\mathcal{K}_v|}\sum_{i\in\mathcal{K}_v}c_v(w,x_i),
&w_v^\star&=\arg\max_{w\in\mathcal{W}}\bar c_v(w),\\
\mathrm{Mism}_v&=\frac{1}{|\mathcal{J}_v|}\sum_{i\in\mathcal{J}_v}
\mathbb{1}[\hat w_{v,i}\neq w_v^\star].
\end{aligned}
\end{equation}
}
Exact ties use the fixed role order fact, rule, verifier. Thus Mism. compares each valid report with the cohort-level top causal contributor rather than selecting a possibly non-unique top role per instance.

\begin{table*}[t]
\centering
\footnotesize
\setlength{\tabcolsep}{5.5pt}
\begin{tabular}{@{}l ccccc@{}}
\toprule
Model & CAP & LR & PR & FPR & Mism. \\
\midrule
\rowcolor{black!5} \textsc{Qwen}-7B  & .34 & .34 & .00 & 1.00 & .50 \\
\rowcolor{black!5} \textsc{Qwen}-14B & .33 & .33 & .01 & 1.00 & .02 \\
\rowcolor{black!5} \textsc{Qwen}-32B & .25 & .20 & .17 & 1.00 & .93 \\
\specialrule{0.5pt}{1.0pt}{1.0pt}
\rowcolor{black!9} \textsc{DSR1}-7B  & .28 & .28 & .00 & .59 & .58 \\
\rowcolor{black!9} \textsc{DSR1}-14B & .39 & .39 & .00 & .97 & .89 \\
\rowcolor{black!9} \textsc{DSR1}-32B & .35 & .33 & .04 & .95 & .03 \\
\specialrule{0.5pt}{1.0pt}{1.0pt}
\rowcolor{black!5} \textsc{GPT-4.1}-nano & .30 & .29 & .03 & 1.00 & .99 \\
\rowcolor{black!5} \textsc{GPT-4.1}-mini & .03 & .03 & .08 & 1.00 & .57 \\
\specialrule{0.5pt}{1.0pt}{1.0pt}
\rowcolor{black!9} \textsc{GPT-5}-nano & .13 & .09 & .31 & .99 & .72 \\
\rowcolor{black!9} \textsc{GPT-5}-mini & .01 & .00 & .71 & 1.00 & .97 \\
\specialrule{0.5pt}{1.0pt}{1.0pt}
\rowcolor{black!5} \textsc{GPT-5.4}-nano & .11 & .11 & .07 & 1.00 & .16 \\
\rowcolor{black!5} \textsc{GPT-5.4}-mini & .20 & .13 & .30 & 1.00 & .69 \\
\bottomrule
\end{tabular}
\caption{Complete Logic multi-agent audit under ordinary messages; Mism. compares attributed and intervention-based top contributors.}
\label{tab:mas-logic-full}
\end{table*}

\paragraph{Laundering across the full cohort.}
Table~\ref{tab:mas-logic-full} shows that the cross-domain subset is not selected for unusually poor provenance. Eleven of twelve Logic systems exhibit skill laundering. For Qwen-7B, Qwen-14B, DSR1-7B, and DSR1-14B, LR equals CAP, so every observed propagation event loses the correct source under ordinary messages. GPT-5-mini gives the complementary failure. PR is $.71$ while CAP is only $.01$. Source recovery and causal propagation are therefore not equivalent properties.

\paragraph{Organization is not a universal amplifier.}
Figure~\ref{fig:mas-organization} compares each distributed system with its model-matched single-agent audit. Logic effects range from $-.47$ to $+.05$. GPT-5.4-nano shows decisive attenuation ($-.47$, 95\% CI $[-.55,-.37]$), while both positive Qwen estimates cross zero. Every Mathematics interval also crosses zero. Multi-agent organization can therefore amplify or attenuate the same skill, but the direction is model- and domain-dependent. This heterogeneity is why the main result concerns provenance loss after measured propagation rather than claiming universal amplification.

\paragraph{Attributed roles need not be causal contributors.}
The final column of Table~\ref{tab:mas-logic-full} compares the most influential role named by the team with agent-removal counterfactuals. Contribution mismatch ranges from $.02$ to $.99$. Qwen-32B and GPT-4.1-nano almost always name a role other than the intervention-based top contributor, whereas Qwen-14B and DSR1-32B are closely aligned. Role descriptions therefore cannot substitute for removal-based contribution estimates, even when a team returns a complete provenance record.

\section{Additional Analysis for RQ1 and RQ4}
\label{app:rq1-rq4}

\paragraph{Detector construction.}
Table~\ref{tab:detectors} treats paired deletion reliance $r_v$ as the target and evaluates signals available from the ordinary skill-conditioned run alone. \emph{Explicit attribution} uses the extracted skill-use bit $a_v$. \emph{Skill mention} requires the phrase \emph{forward chain} in the visible reasoning trace. \emph{Trace--skill similarity} compares embeddings of the trace and skill body using a per-model median threshold. \emph{LLM judge} receives the skill and visible reasoning trace and predicts whether the procedure affected the answer. Precision measures how often a positive detector prediction coincides with $r_v=1$; recall measures how many reliance events it recovers. The target prevalence is $.34$, so precision near this value indicates little discrimination despite apparently high recall.

\paragraph{Why outcome improvement is not an observational detector.}
For reference, labeling only wrong-to-correct flips as skill use gives precision $1.00$, recall $.61$, F1 $.76$, and a positive rate of $.20$. This construction is precise by definition: it compares the skill-conditioned decision with both the no-skill decision and the gold answer. It is therefore an intervention-based diagnostic rather than a cheaper alternative to \backtrace{}. Moreover, it omits correct-to-wrong and same-correctness answer changes, both of which remain evidence that the artifact influenced the decision. The comparison shows why neither accuracy gain nor a detector built from one visible run can replace the paired reliance target.

\paragraph{Attribution-position control.}
The main protocol commits an answer before requesting attribution so that the attribution query cannot alter the decision it describes. As a control, we instead elicit the answer and attribution in one generation. Table~\ref{tab:audit-ablations} shows that attribution remains $.98$--$1.00$ across the twelve models, while conditional performative-use rates differ from the separated protocol by less than $.02$ at the median and at most $.10$. The pervasive positive attribution signal is therefore not an artifact of asking a second question after the answer.

\begin{table*}[t]
\centering
\footnotesize
\setlength{\tabcolsep}{5.0pt}
\begin{tabular}{@{}l cc cc cc@{}}
\toprule
& \multicolumn{2}{c}{Integrated attribution} & \multicolumn{2}{c}{CAP} & \multicolumn{2}{c}{PR} \\
\cmidrule(lr){2-3} \cmidrule(lr){4-5} \cmidrule(l){6-7}
Model & $A$ & PUR & Asym. & Redun. & Asym. & Redun. \\
\midrule
\rowcolor{black!5} \textsc{Qwen}-7B  & 1.00 & .70 & .34 & .47 & .00 & .01 \\
\rowcolor{black!5} \textsc{Qwen}-14B & 1.00 & .73 & .33 & .33 & .01 & .03 \\
\rowcolor{black!5} \textsc{Qwen}-32B & 1.00 & .80 & .25 & .23 & .17 & .10 \\
\specialrule{0.5pt}{1.0pt}{1.0pt}
\rowcolor{black!9} \textsc{DSR1}-7B  & 1.00 & .53 & .28 & .56 & .00 & .00 \\
\rowcolor{black!9} \textsc{DSR1}-14B & 1.00 & .59 & .39 & .43 & .00 & .07 \\
\rowcolor{black!9} \textsc{DSR1}-32B & .98 & .70 & .35 & .38 & .04 & .17 \\
\specialrule{0.5pt}{1.0pt}{1.0pt}
\rowcolor{black!5} \textsc{GPT-4.1}-nano & .99 & .65 & .30 & .39 & .03 & .25 \\
\rowcolor{black!5} \textsc{GPT-4.1}-mini & 1.00 & .93 & .03 & .02 & .08 & .18 \\
\specialrule{0.5pt}{1.0pt}{1.0pt}
\rowcolor{black!9} \textsc{GPT-5}-nano & 1.00 & .85 & .13 & .17 & .31 & .19 \\
\rowcolor{black!9} \textsc{GPT-5}-mini & 1.00 & 1.00 & .01 & .01 & .71 & .60 \\
\specialrule{0.5pt}{1.0pt}{1.0pt}
\rowcolor{black!5} \textsc{GPT-5.4}-nano & 1.00 & .45 & .11 & .15 & .07 & .01 \\
\rowcolor{black!5} \textsc{GPT-5.4}-mini & 1.00 & .45 & .20 & .14 & .30 & .43 \\
\bottomrule
\end{tabular}
\caption{Attribution-position and skill-assignment ablations on Logic.}
\label{tab:audit-ablations}
\end{table*}

\paragraph{Redundant-assignment control.}
The same table compares the primary asymmetric organization, in which one specialist holds the skill, with a redundant organization that supplies it to every specialist. Redundancy can increase or decrease CAP, and it does not consistently recover provenance. DSR1-7B doubles CAP from $.28$ to $.56$ while PR remains zero; Qwen-32B and GPT-5-nano instead lose provenance despite similar CAP. Provenance loss is therefore not an artifact of hiding the skill from most team members.

\paragraph{Saturated-performance control.}
We repeat the audit on 200 shallow Logic instances without adversarial traps. The skill library, intervention conditions, normalization, and attribution extraction remain unchanged. Table~\ref{tab:rq4-saturated} reports the main twelve-model cohort. Eight models reach $1.00$ baseline and correct-skill accuracy; for all eight, deletion changes no answer, yet their performative-use rate is $1.00$. The irrelevant-skill attribution rate simultaneously ranges from zero to one. Attribution failure therefore persists when ordinary task error has largely disappeared and cannot be reduced to confusion on difficult instances.

\begin{table*}[htbp]
\centering
\footnotesize
\setlength{\tabcolsep}{3.0pt}
\begin{tabular}{@{}lccccc@{}}
\toprule
& & \multicolumn{3}{c}{Correct} & Irrel. \\
\cmidrule(lr){3-5} \cmidrule(l){6-6}
Model & Base & Acc & $R$ & PUR & $A$ \\
\midrule
\rowcolor{black!5} \textsc{Qwen}-7B  & 1.00 & 1.00 & .00 & 1.00 & .94 \\
\rowcolor{black!5} \textsc{Qwen}-14B & 1.00 & 1.00 & .00 & 1.00 & .00 \\
\rowcolor{black!5} \textsc{Qwen}-32B & 1.00 & 1.00 & .00 & 1.00 & .18 \\
\specialrule{0.5pt}{1.0pt}{1.0pt}
\rowcolor{black!9} \textsc{DSR1}-7B  & .92 & .93 & .10 & .91 & .66 \\
\rowcolor{black!9} \textsc{DSR1}-14B & .99 & .99 & .00 & 1.00 & .10 \\
\rowcolor{black!9} \textsc{DSR1}-32B & 1.00 & 1.00 & .00 & 1.00 & .56 \\
\specialrule{0.5pt}{1.0pt}{1.0pt}
\rowcolor{black!5} \textsc{GPT-4.1}-nano & 1.00 & .99 & .01 & .99 & 1.00 \\
\rowcolor{black!5} \textsc{GPT-4.1}-mini & 1.00 & 1.00 & .00 & 1.00 & .04 \\
\specialrule{0.5pt}{1.0pt}{1.0pt}
\rowcolor{black!9} \textsc{GPT-5}-nano & 1.00 & 1.00 & .00 & 1.00 & .42 \\
\rowcolor{black!9} \textsc{GPT-5}-mini & 1.00 & 1.00 & .00 & 1.00 & .00 \\
\specialrule{0.5pt}{1.0pt}{1.0pt}
\rowcolor{black!5} \textsc{GPT-5.4}-nano & 1.00 & .99 & .01 & .99 & .51 \\
\rowcolor{black!5} \textsc{GPT-5.4}-mini & 1.00 & 1.00 & .00 & 1.00 & 1.00 \\
\bottomrule
\end{tabular}
\caption{Saturated-performance control on 200 shallow Logic instances.}
\label{tab:rq4-saturated}
\end{table*}

\section{Prompt and Output Schemas}
\label{app:schemas}

\paragraph{Decision channel.}
Logic prompts request a derivation followed by exactly \texttt{Answer: True} or \texttt{Answer: False}. Mathematics prompts request a final \texttt{\string\boxed\{...\}} expression. The Logic normalizer maps only the terminal Boolean field, while the Mathematics normalizer canonicalizes equivalent numeric and symbolic expressions before matching. Reasoning text is retained for observational baselines but never determines the intervention target.

\paragraph{Attribution channel.}
After storing the decision, a separate turn lists the exact display names available in that condition and requests the schema
\begin{quote}\small\ttfamily
\{"used\_skills": ["..."], "unused\_skills": ["..."]\}.
\end{quote}
The request forbids re-solving or revising the answer. Names are matched case-insensitively against the supplied artifact list; an unknown name, a missing required field, or a non-list value fails validation. The None condition uses the same schema with an empty supplied list, so any named artifact is a false-attribution event.

\paragraph{Distributed provenance.}
The team-level schema additionally records the most influential agent, the skills used in the pipeline, and the agent that supplied each named skill. Full provenance recovery requires both the correct artifact and its designated source. Naming the artifact without the source does not count as recovery. All fields are elicited only after the aggregator commits the team decision.

\paragraph{Repair and missing outputs.}
A malformed attribution record receives at most one format-only repair request containing the invalid text and target schema; the request explicitly forbids new reasoning or answer revision. If repair fails, the terminal generation remains available for coverage accounting. Accuracy treats an unresolved decision as incorrect, while paired causal and attribution metrics omit records lacking either a normalized decision or a valid attribution object.

\section{Validity and Scope}
\label{app:scope}

\paragraph{What the intervention identifies.}
\backtrace{} identifies whether changing a supplied artifact changes a normalized decision under a fixed execution context. It does not expose private token-level computation, and an unchanged answer cannot rule out use of a similar procedure learned during pretraining. The conclusions therefore concern artifact-level causal sensitivity and its observable attribution, not a complete reconstruction of internal reasoning.

\paragraph{Domain and decoding boundaries.}
The Logic track guarantees verified proofs and decision-relevant traps but remains synthetic. Mathematics supplies heterogeneous human-authored problems, yet no construction guarantees that every problem requires the provided procedure. Greedy local decoding and temperature-zero API calls where supported measure one execution path per condition and do not estimate serving variance. Near-chance regimes may also amplify sensitivity to small prompt changes, which is why directional interpretations use helpful and harmful flips rather than reliance alone.

\paragraph{Conditional metrics and coverage.}
SUR is defined only when reliance events exist, and PUR only when attribution events exist; an empty denominator is a structural property rather than evidence of fidelity. Accuracy retains every terminal generation, whereas $R$, $A$, $\Gamma$, SUR, PUR, and AFS use valid paired decisions and attribution objects. Effective coverage and parse rates are exposed rather than imputing missing observations.

\section{Complete Single-Agent Results}
\label{app:complete-results}

The main tables retain the compact \(\Delta\mathrm{Acc}\)--AFS view needed for comparison. The following tables provide every aggregate metric for all model--condition pairs. \(R\) and \(A\) are descriptive rates; \(\Gamma\), SUR, and PUR quantify attribution error; AFS measures overlap between reliance and attribution. A dash marks a conditional rate whose denominator is empty.

\paragraph{AFS under saturated attribution.}
When every instance is attributed and SUR is zero, AFS reduces algebraically to $2R/(1+R)$. This is the limiting case of indiscriminate attribution: the score reports overlap with the varying reliance set rather than supplying a second causal signal. It does not characterize the full evaluation. Across the 108 model--condition cells below, 58 have nonzero SUR, including 17 of 18 Irrelevant, 14 of 18 Content swap, and 10 of 18 Name swap cells. AFS is therefore a compact agreement summary; causal evidence remains the paired reliance variable that it compares with attribution.

\begin{table*}[t]
\centering
\footnotesize
\setlength{\tabcolsep}{4.2pt}
\begin{tabular}{@{}lcccccccc@{}}
\toprule
\multicolumn{9}{c}{Logic: Correct} \\
Model & Acc & $\Delta$Acc & $R$ & $A$ & $\Gamma$ & SUR & PUR & AFS \\
\midrule
\rowcolor{black!5} \textsc{Qwen}-7B & .43 & -.05 & .29 & 1.00 & .71 & .00 & .71 & .45 \\
\rowcolor{black!5} \textsc{Qwen}-14B & .74 & +.17 & .29 & 1.00 & .71 & .00 & .71 & .45 \\
\rowcolor{black!5} \textsc{Qwen}-32B & .79 & .00 & .21 & 1.00 & .79 & .00 & .79 & .35 \\
\rowcolor{black!9} \textsc{DSR1}-7B & .47 & -.01 & .38 & .98 & .62 & .04 & .62 & .54 \\
\rowcolor{black!9} \textsc{DSR1}-14B & .68 & .00 & .33 & .99 & .66 & .00 & .67 & .50 \\
\rowcolor{black!9} \textsc{DSR1}-32B & .73 & +.03 & .20 & 1.00 & .80 & .00 & .80 & .33 \\
\rowcolor{black!5} \textsc{GPT-4.1}-nano & .59 & .00 & .38 & .99 & .61 & .00 & .62 & .55 \\
\rowcolor{black!5} \textsc{GPT-4.1}-mini & .95 & +.02 & .09 & 1.00 & .92 & .04 & .92 & .15 \\
\rowcolor{black!9} \textsc{GPT-5}-nano & .85 & -.01 & .19 & .97 & .82 & .11 & .82 & .30 \\
\rowcolor{black!9} \textsc{GPT-5}-mini & .99 & -.01 & .01 & 1.00 & .99 & .00 & .99 & .01 \\
\rowcolor{black!5} \textsc{GPT-5.4}-nano & .88 & +.49 & .56 & 1.00 & .44 & .00 & .44 & .72 \\
\rowcolor{black!5} \textsc{GPT-5.4}-mini & .81 & +.52 & .56 & 1.00 & .44 & .00 & .44 & .72 \\
\bottomrule
\end{tabular}
\caption{Complete Logic results for the correct condition.}
\label{tab:full-logic-correct}
\end{table*}

\begin{table*}[t]
\centering
\footnotesize
\setlength{\tabcolsep}{4.2pt}
\begin{tabular}{@{}lcccccccc@{}}
\toprule
\multicolumn{9}{c}{Logic: Paraphrase} \\
Model & Acc & $\Delta$Acc & $R$ & $A$ & $\Gamma$ & SUR & PUR & AFS \\
\midrule
\rowcolor{black!5} \textsc{Qwen}-7B & .46 & -.03 & .36 & 1.00 & .64 & .00 & .64 & .53 \\
\rowcolor{black!5} \textsc{Qwen}-14B & .71 & +.14 & .32 & 1.00 & .68 & .00 & .68 & .49 \\
\rowcolor{black!5} \textsc{Qwen}-32B & .81 & +.03 & .20 & 1.00 & .80 & .00 & .80 & .33 \\
\rowcolor{black!9} \textsc{DSR1}-7B & .49 & +.02 & .36 & .96 & .63 & .04 & .64 & .53 \\
\rowcolor{black!9} \textsc{DSR1}-14B & .73 & +.04 & .31 & 1.00 & .69 & .00 & .69 & .48 \\
\rowcolor{black!9} \textsc{DSR1}-32B & .76 & +.05 & .18 & 1.00 & .82 & .00 & .82 & .31 \\
\rowcolor{black!5} \textsc{GPT-4.1}-nano & .60 & +.02 & .37 & 1.00 & .63 & .00 & .63 & .54 \\
\rowcolor{black!5} \textsc{GPT-4.1}-mini & .96 & +.02 & .07 & 1.00 & .93 & .00 & .93 & .14 \\
\rowcolor{black!9} \textsc{GPT-5}-nano & .82 & -.05 & .23 & .96 & .78 & .13 & .79 & .34 \\
\rowcolor{black!9} \textsc{GPT-5}-mini & .99 & -.01 & .01 & 1.00 & .99 & .00 & .99 & .02 \\
\rowcolor{black!5} \textsc{GPT-5.4}-nano & .86 & +.47 & .53 & 1.00 & .47 & .00 & .47 & .69 \\
\rowcolor{black!5} \textsc{GPT-5.4}-mini & .84 & +.56 & .58 & 1.00 & .42 & .00 & .42 & .74 \\
\bottomrule
\end{tabular}
\caption{Complete Logic results for the paraphrase condition.}
\label{tab:full-logic-paraphrase}
\end{table*}

\begin{table*}[t]
\centering
\footnotesize
\setlength{\tabcolsep}{4.2pt}
\begin{tabular}{@{}lcccccccc@{}}
\toprule
\multicolumn{9}{c}{Logic: Misleading} \\
Model & Acc & $\Delta$Acc & $R$ & $A$ & $\Gamma$ & SUR & PUR & AFS \\
\midrule
\rowcolor{black!5} \textsc{Qwen}-7B & .47 & -.01 & .33 & 1.00 & .67 & .00 & .67 & .50 \\
\rowcolor{black!5} \textsc{Qwen}-14B & .72 & +.15 & .30 & 1.00 & .70 & .00 & .70 & .47 \\
\rowcolor{black!5} \textsc{Qwen}-32B & .79 & .00 & .22 & 1.00 & .78 & .00 & .78 & .36 \\
\rowcolor{black!9} \textsc{DSR1}-7B & .48 & +.01 & .37 & .98 & .62 & .02 & .62 & .54 \\
\rowcolor{black!9} \textsc{DSR1}-14B & .68 & .00 & .34 & 1.00 & .66 & .00 & .66 & .51 \\
\rowcolor{black!9} \textsc{DSR1}-32B & .72 & +.02 & .15 & 1.00 & .85 & .00 & .85 & .26 \\
\rowcolor{black!5} \textsc{GPT-4.1}-nano & .57 & -.01 & .34 & 1.00 & .66 & .00 & .66 & .50 \\
\rowcolor{black!5} \textsc{GPT-4.1}-mini & .97 & +.03 & .06 & .99 & .94 & .06 & .94 & .11 \\
\rowcolor{black!9} \textsc{GPT-5}-nano & .85 & -.01 & .22 & .98 & .79 & .08 & .79 & .34 \\
\rowcolor{black!9} \textsc{GPT-5}-mini & .81 & -.19 & .19 & 1.00 & .81 & .02 & .81 & .31 \\
\rowcolor{black!5} \textsc{GPT-5.4}-nano & .86 & +.46 & .52 & 1.00 & .48 & .00 & .48 & .69 \\
\rowcolor{black!5} \textsc{GPT-5.4}-mini & .78 & +.50 & .54 & 1.00 & .46 & .00 & .46 & .70 \\
\bottomrule
\end{tabular}
\caption{Complete Logic results for the misleading condition.}
\label{tab:full-logic-misleading}
\end{table*}

\begin{table*}[t]
\centering
\footnotesize
\setlength{\tabcolsep}{4.2pt}
\begin{tabular}{@{}lcccccccc@{}}
\toprule
\multicolumn{9}{c}{Logic: Name swap} \\
Model & Acc & $\Delta$Acc & $R$ & $A$ & $\Gamma$ & SUR & PUR & AFS \\
\midrule
\rowcolor{black!5} \textsc{Qwen}-7B & .50 & +.02 & .34 & .98 & .65 & .01 & .66 & .51 \\
\rowcolor{black!5} \textsc{Qwen}-14B & .73 & +.16 & .31 & .71 & .56 & .26 & .68 & .45 \\
\rowcolor{black!5} \textsc{Qwen}-32B & .82 & +.03 & .21 & 1.00 & .79 & .00 & .79 & .34 \\
\rowcolor{black!9} \textsc{DSR1}-7B & .44 & -.03 & .37 & .65 & .56 & .37 & .65 & .45 \\
\rowcolor{black!9} \textsc{DSR1}-14B & .70 & +.01 & .36 & .88 & .60 & .11 & .63 & .52 \\
\rowcolor{black!9} \textsc{DSR1}-32B & .75 & +.04 & .15 & .94 & .80 & .03 & .85 & .26 \\
\rowcolor{black!5} \textsc{GPT-4.1}-nano & .59 & +.01 & .34 & .87 & .62 & .14 & .67 & .48 \\
\rowcolor{black!5} \textsc{GPT-4.1}-mini & .96 & +.02 & .08 & 1.00 & .92 & .00 & .92 & .15 \\
\rowcolor{black!9} \textsc{GPT-5}-nano & .86 & -.01 & .21 & .86 & .73 & .18 & .80 & .32 \\
\rowcolor{black!9} \textsc{GPT-5}-mini & .99 & -.01 & .01 & 1.00 & .99 & .00 & .99 & .02 \\
\rowcolor{black!5} \textsc{GPT-5.4}-nano & .82 & +.43 & .48 & 1.00 & .52 & .00 & .52 & .65 \\
\rowcolor{black!5} \textsc{GPT-5.4}-mini & .79 & +.50 & .53 & 1.00 & .47 & .00 & .47 & .69 \\
\bottomrule
\end{tabular}
\caption{Complete Logic results for the name swap condition.}
\label{tab:full-logic-name-swap}
\end{table*}

\begin{table*}[t]
\centering
\footnotesize
\setlength{\tabcolsep}{4.2pt}
\begin{tabular}{@{}lcccccccc@{}}
\toprule
\multicolumn{9}{c}{Logic: Content swap} \\
Model & Acc & $\Delta$Acc & $R$ & $A$ & $\Gamma$ & SUR & PUR & AFS \\
\midrule
\rowcolor{black!5} \textsc{Qwen}-7B & .47 & -.01 & .32 & .96 & .68 & .05 & .69 & .47 \\
\rowcolor{black!5} \textsc{Qwen}-14B & .58 & +.01 & .26 & .97 & .76 & .08 & .76 & .38 \\
\rowcolor{black!5} \textsc{Qwen}-32B & .78 & -.01 & .18 & 1.00 & .82 & .02 & .82 & .30 \\
\rowcolor{black!9} \textsc{DSR1}-7B & .48 & +.01 & .40 & .79 & .56 & .21 & .60 & .53 \\
\rowcolor{black!9} \textsc{DSR1}-14B & .64 & -.04 & .32 & .30 & .39 & .64 & .61 & .37 \\
\rowcolor{black!9} \textsc{DSR1}-32B & .72 & +.01 & .16 & .65 & .58 & .28 & .82 & .29 \\
\rowcolor{black!5} \textsc{GPT-4.1}-nano & .55 & -.03 & .40 & .99 & .60 & .00 & .60 & .57 \\
\rowcolor{black!5} \textsc{GPT-4.1}-mini & .93 & .00 & .09 & .97 & .92 & .19 & .93 & .13 \\
\rowcolor{black!9} \textsc{GPT-5}-nano & .85 & -.02 & .20 & .63 & .65 & .56 & .86 & .22 \\
\rowcolor{black!9} \textsc{GPT-5}-mini & 1.00 & .00 & .00 & 1.00 & 1.00 & .00 & 1.00 & .01 \\
\rowcolor{black!5} \textsc{GPT-5.4}-nano & .46 & +.07 & .11 & .85 & .79 & .19 & .90 & .18 \\
\rowcolor{black!5} \textsc{GPT-5.4}-mini & .28 & -.01 & .14 & 1.00 & .86 & .00 & .86 & .24 \\
\bottomrule
\end{tabular}
\caption{Complete Logic results for the content swap condition.}
\label{tab:full-logic-content-swap}
\end{table*}

\begin{table*}[t]
\centering
\footnotesize
\setlength{\tabcolsep}{4.2pt}
\begin{tabular}{@{}lcccccccc@{}}
\toprule
\multicolumn{9}{c}{Logic: Irrelevant} \\
Model & Acc & $\Delta$Acc & $R$ & $A$ & $\Gamma$ & SUR & PUR & AFS \\
\midrule
\rowcolor{black!5} \textsc{Qwen}-7B & .46 & -.02 & .44 & .69 & .53 & .32 & .56 & .53 \\
\rowcolor{black!5} \textsc{Qwen}-14B & .58 & +.01 & .31 & .00 & .31 & 1.00 & -- & .00 \\
\rowcolor{black!5} \textsc{Qwen}-32B & .76 & -.02 & .25 & .23 & .35 & .73 & .71 & .28 \\
\rowcolor{black!9} \textsc{DSR1}-7B & .52 & +.05 & .38 & .68 & .59 & .39 & .66 & .44 \\
\rowcolor{black!9} \textsc{DSR1}-14B & .58 & -.10 & .35 & .19 & .38 & .78 & .59 & .29 \\
\rowcolor{black!9} \textsc{DSR1}-32B & .68 & -.03 & .17 & .89 & .73 & .06 & .82 & .31 \\
\rowcolor{black!5} \textsc{GPT-4.1}-nano & .52 & -.07 & .37 & .96 & .64 & .06 & .64 & .52 \\
\rowcolor{black!5} \textsc{GPT-4.1}-mini & .92 & -.02 & .11 & .09 & .20 & .97 & .96 & .03 \\
\rowcolor{black!9} \textsc{GPT-5}-nano & .87 & +.01 & .19 & .23 & .34 & .78 & .81 & .20 \\
\rowcolor{black!9} \textsc{GPT-5}-mini & .99 & -.01 & .01 & .00 & .01 & 1.00 & -- & .00 \\
\rowcolor{black!5} \textsc{GPT-5.4}-nano & .42 & +.02 & .10 & .95 & .87 & .10 & .91 & .17 \\
\rowcolor{black!5} \textsc{GPT-5.4}-mini & .28 & .00 & .10 & 1.00 & .90 & .00 & .90 & .19 \\
\bottomrule
\end{tabular}
\caption{Complete Logic results for the irrelevant condition.}
\label{tab:full-logic-irrelevant}
\end{table*}

\begin{table*}[t]
\centering
\footnotesize
\setlength{\tabcolsep}{4.2pt}
\begin{tabular}{@{}lcccccccc@{}}
\toprule
\multicolumn{9}{c}{Mathematics: Correct} \\
Model & Acc & $\Delta$Acc & $R$ & $A$ & $\Gamma$ & SUR & PUR & AFS \\
\midrule
\rowcolor{black!5} \textsc{Qwen}-14B & .78 & .00 & .15 & .98 & .83 & .00 & .85 & .26 \\
\rowcolor{black!5} \textsc{Qwen}-32B & .76 & -.02 & .15 & .96 & .81 & .00 & .84 & .27 \\
\rowcolor{black!9} \textsc{DSR1}-14B & .87 & +.03 & .10 & .87 & .78 & .04 & .89 & .20 \\
\rowcolor{black!5} \textsc{GPT-4.1}-mini & .86 & -.01 & .09 & .99 & .90 & .00 & .91 & .17 \\
\rowcolor{black!9} \textsc{GPT-5}-mini & .88 & .00 & .07 & .99 & .92 & .00 & .93 & .14 \\
\rowcolor{black!5} \textsc{GPT-5.4}-nano & .79 & +.01 & .20 & 1.00 & .81 & .02 & .81 & .33 \\
\bottomrule
\end{tabular}
\caption{Complete Mathematics results for the correct condition.}
\label{tab:full-math-correct}
\end{table*}

\begin{table*}[t]
\centering
\footnotesize
\setlength{\tabcolsep}{4.2pt}
\begin{tabular}{@{}lcccccccc@{}}
\toprule
\multicolumn{9}{c}{Mathematics: Paraphrase} \\
Model & Acc & $\Delta$Acc & $R$ & $A$ & $\Gamma$ & SUR & PUR & AFS \\
\midrule
\rowcolor{black!5} \textsc{Qwen}-14B & .77 & -.01 & .17 & .99 & .82 & .00 & .83 & .29 \\
\rowcolor{black!5} \textsc{Qwen}-32B & .77 & -.02 & .14 & .95 & .82 & .03 & .85 & .25 \\
\rowcolor{black!9} \textsc{DSR1}-14B & .87 & +.03 & .09 & .80 & .73 & .08 & .90 & .18 \\
\rowcolor{black!5} \textsc{GPT-4.1}-mini & .87 & .00 & .10 & .98 & .90 & .11 & .91 & .16 \\
\rowcolor{black!9} \textsc{GPT-5}-mini & .87 & -.02 & .07 & .99 & .94 & .11 & .94 & .11 \\
\rowcolor{black!5} \textsc{GPT-5.4}-nano & .82 & +.04 & .16 & 1.00 & .84 & .00 & .84 & .27 \\
\bottomrule
\end{tabular}
\caption{Complete Mathematics results for the paraphrase condition.}
\label{tab:full-math-paraphrase}
\end{table*}

\begin{table*}[t]
\centering
\footnotesize
\setlength{\tabcolsep}{4.2pt}
\begin{tabular}{@{}lcccccccc@{}}
\toprule
\multicolumn{9}{c}{Mathematics: Misleading} \\
Model & Acc & $\Delta$Acc & $R$ & $A$ & $\Gamma$ & SUR & PUR & AFS \\
\midrule
\rowcolor{black!5} \textsc{Qwen}-14B & .76 & -.02 & .18 & .98 & .80 & .00 & .81 & .32 \\
\rowcolor{black!5} \textsc{Qwen}-32B & .80 & +.02 & .17 & .95 & .79 & .02 & .83 & .29 \\
\rowcolor{black!9} \textsc{DSR1}-14B & .89 & +.05 & .07 & .89 & .83 & .10 & .93 & .14 \\
\rowcolor{black!5} \textsc{GPT-4.1}-mini & .86 & .00 & .07 & .99 & .92 & .00 & .93 & .14 \\
\rowcolor{black!9} \textsc{GPT-5}-mini & .88 & .00 & .06 & .99 & .93 & .00 & .94 & .11 \\
\rowcolor{black!5} \textsc{GPT-5.4}-nano & .79 & +.01 & .18 & 1.00 & .82 & .00 & .82 & .31 \\
\bottomrule
\end{tabular}
\caption{Complete Mathematics results for the misleading condition.}
\label{tab:full-math-misleading}
\end{table*}

\begin{table*}[t]
\centering
\footnotesize
\setlength{\tabcolsep}{4.2pt}
\begin{tabular}{@{}lcccccccc@{}}
\toprule
\multicolumn{9}{c}{Mathematics: Name swap} \\
Model & Acc & $\Delta$Acc & $R$ & $A$ & $\Gamma$ & SUR & PUR & AFS \\
\midrule
\rowcolor{black!5} \textsc{Qwen}-14B & .76 & -.02 & .16 & .93 & .78 & .04 & .83 & .29 \\
\rowcolor{black!5} \textsc{Qwen}-32B & .78 & .00 & .17 & .99 & .82 & .00 & .83 & .29 \\
\rowcolor{black!9} \textsc{DSR1}-14B & .86 & +.02 & .08 & .88 & .80 & .00 & .91 & .17 \\
\rowcolor{black!5} \textsc{GPT-4.1}-mini & .85 & -.01 & .07 & .97 & .93 & .20 & .94 & .11 \\
\rowcolor{black!9} \textsc{GPT-5}-mini & .88 & +.01 & .07 & .99 & .92 & .00 & .93 & .14 \\
\rowcolor{black!5} \textsc{GPT-5.4}-nano & .83 & +.05 & .16 & .99 & .85 & .04 & .85 & .26 \\
\bottomrule
\end{tabular}
\caption{Complete Mathematics results for the name swap condition.}
\label{tab:full-math-name-swap}
\end{table*}

\begin{table*}[t]
\centering
\footnotesize
\setlength{\tabcolsep}{4.2pt}
\begin{tabular}{@{}lcccccccc@{}}
\toprule
\multicolumn{9}{c}{Mathematics: Content swap} \\
Model & Acc & $\Delta$Acc & $R$ & $A$ & $\Gamma$ & SUR & PUR & AFS \\
\midrule
\rowcolor{black!5} \textsc{Qwen}-14B & .78 & .00 & .17 & .74 & .70 & .36 & .86 & .23 \\
\rowcolor{black!5} \textsc{Qwen}-32B & .77 & .00 & .14 & .93 & .84 & .17 & .87 & .22 \\
\rowcolor{black!9} \textsc{DSR1}-14B & .84 & .00 & .09 & .57 & .54 & .33 & .90 & .18 \\
\rowcolor{black!5} \textsc{GPT-4.1}-mini & .87 & .00 & .07 & .95 & .91 & .16 & .94 & .11 \\
\rowcolor{black!9} \textsc{GPT-5}-mini & .88 & +.01 & .06 & .99 & .93 & .06 & .94 & .11 \\
\rowcolor{black!5} \textsc{GPT-5.4}-nano & .80 & +.02 & .16 & 1.00 & .84 & .00 & .84 & .27 \\
\bottomrule
\end{tabular}
\caption{Complete Mathematics results for the content swap condition.}
\label{tab:full-math-content-swap}
\end{table*}

\begin{table*}[t]
\centering
\footnotesize
\setlength{\tabcolsep}{4.2pt}
\begin{tabular}{@{}lcccccccc@{}}
\toprule
\multicolumn{9}{c}{Mathematics: Irrelevant} \\
Model & Acc & $\Delta$Acc & $R$ & $A$ & $\Gamma$ & SUR & PUR & AFS \\
\midrule
\rowcolor{black!5} \textsc{Qwen}-14B & .77 & -.01 & .18 & .00 & .18 & 1.00 & -- & .00 \\
\rowcolor{black!5} \textsc{Qwen}-32B & .78 & .00 & .13 & .00 & .13 & 1.00 & -- & .00 \\
\rowcolor{black!9} \textsc{DSR1}-14B & .87 & +.03 & .11 & .55 & .52 & .35 & .87 & .22 \\
\rowcolor{black!5} \textsc{GPT-4.1}-mini & .86 & .00 & .07 & .00 & .07 & 1.00 & -- & .00 \\
\rowcolor{black!9} \textsc{GPT-5}-mini & .87 & -.01 & .10 & .13 & .19 & .78 & .84 & .18 \\
\rowcolor{black!5} \textsc{GPT-5.4}-nano & .80 & +.02 & .19 & .49 & .39 & .23 & .70 & .43 \\
\bottomrule
\end{tabular}
\caption{Complete Mathematics results for the irrelevant condition.}
\label{tab:full-math-irrelevant}
\end{table*}

\end{document}